\documentclass{article}

\PassOptionsToPackage{numbers, compress}{natbib}
\usepackage[main, final]{neurips_2026}

\usepackage[utf8]{inputenc}
\usepackage[T1]{fontenc}
\usepackage{xcolor}
\usepackage{hyperref}
\usepackage{url}
\usepackage{booktabs}
\usepackage{amsfonts}
\usepackage{nicefrac}
\usepackage{microtype}
\usepackage{graphicx}
\usepackage{caption}
\usepackage{amsmath}
\usepackage{tikz}
\usepackage{natbib}
\usepackage{enumitem}
\usepackage{seqsplit}
\usepackage{subcaption}
\definecolor{lightgreen}{RGB}{220, 240, 210}
\definecolor{citeblue}{rgb}{0.03,0.1,0.4}
\definecolor{linkred}{rgb}{0.9,0,0}

\usepackage{colortbl}
\usepackage{arydshln}
\usepackage{dashrule}
\usepackage{wrapfig}
\definecolor{gg}{gray}{0.92}
\newcolumntype{a}{>{\columncolor{gg}}c}
\usepackage{tcolorbox}

\hypersetup{
    colorlinks = true,
    citecolor  = citeblue,
    linkcolor  = linkred,
    urlcolor = citeblue
}
\tcbuselibrary{skins, breakable}
\title{ES-Merging: Biological MLLM Merging \\via Embedding Space Signals}

\author{Wonbin Lee\thanks{Equal Contribution}$^{\ast 1}$ \quad Dongki Kim$^{\ast 1}$ \quad Sung Ju Hwang$^{1,2}$ \\[0.3em]
    KAIST$^{1}$, DeepAuto.ai$^{2}$ \\[0.3em]
    \texttt{\{smilelwb01, cleverki, sungju.hwang\}@kaist.ac.kr}
}

\begin{document}

\maketitle
\begin{abstract}
Biological multimodal large language models (MLLMs) have emerged as powerful foundation models for scientific discovery. However, existing models are specialized to a single modality, limiting their ability to solve inherently cross-modal scientific problems. While model merging is an efficient method to combine the different modalities into a unified MLLM, existing methods rely on input-agnostic parameter space heuristics that fail to faithfully capture modality specialization. To overcome this limitation, we propose the \textbf{E}mbedding-\textbf{S}ignal-based MLLM \textbf{Merging} (\textbf{ES-Merging}), a framework that estimates merging coefficients from embedding space signals, moving the merging paradigm from the parameter signals to the embedding signals.
ES-Merging exploits coarse-grained and fine-grained signals from embedding space to estimate the layer-wise and element-wise merging coefficients, respectively, which are jointly combined for complementary coefficient estimation. Through extensive experiments, we demonstrate that ES-Merging outperforms existing merging methods not only on the cross-modal reasoning but also on the single-modal knowledge preservation, establishing that embedding space signals provide a principled and effective foundation for MLLM merging. Our code is available at \url{https://github.com/WonbinLee067/es-merging}.
\end{abstract}
\section{Introduction}
Multimodal Large Language Models (MLLMs) have been emerging as crucial foundational models for scientific discovery, extending their perception to diverse biological modalities across molecules~\citep{llamo,molllama}, proteins~\citep{prot2text,prot2textv2}, and cells~\citep{cello1,c2sscale}.
Despite their impressive progress within each modality, diverse scientific problems of interest are inherently cross-modal, requiring to understand the interactive effects such as protein-ligand interactions or drug effectiveness to cell types. However, existing biological MLLMs are specialized to a single modality, resulting in limited intersectional knowledge across diverse biological modalities and unreliable cross-modal reasoning.

Although model merging has been actively explored as an efficient scheme for integrating multiple models, how to merge modality-specialized MLLMs remains largely underexplored. In this setting, the central challenge is how to accurately measure modality specialization of an MLLM that induces the modality-specific behavior. Although modality specialization is ultimately encoded in model parameters, its functional effect is revealed through the representation transformations induced by modality-specific inputs. However, existing merging methods typically rely on input-blind parameter-space heuristics, such as magnitudes, signs, or directions of parameter updates. These signals can provide useful but indirect signals of specialization, since they do not observe how each expert responds to modality token embeddings. This motivates estimating merging coefficients from embedding-space signals that directly reflect input-conditioned modality behavior.

\begin{figure*}[t]
    \begin{minipage}[!]{0.435\linewidth}
        \centering
        \includegraphics[width=0.99\linewidth]{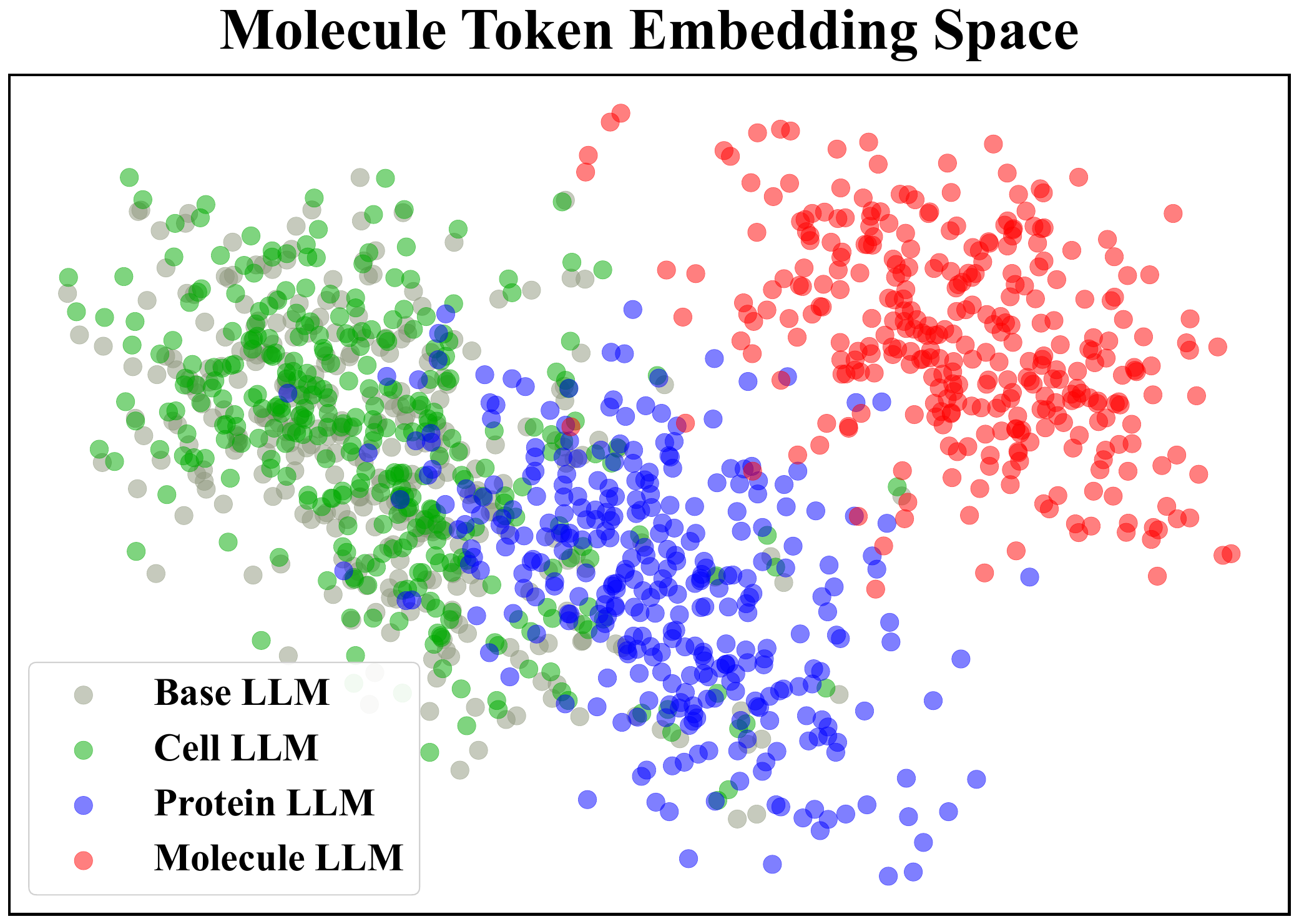}
        \vspace{-0.1in}
        \captionof{figure}{Molecule token embedding visualization of the last transformer block for the base LLM and each specialized LLM.}
        \label{fig:mol_embedding_visualization}
    \end{minipage}
    \hfill
    \begin{minipage}[!]{0.545\linewidth}
        \centering
        \includegraphics[width=0.99\linewidth]{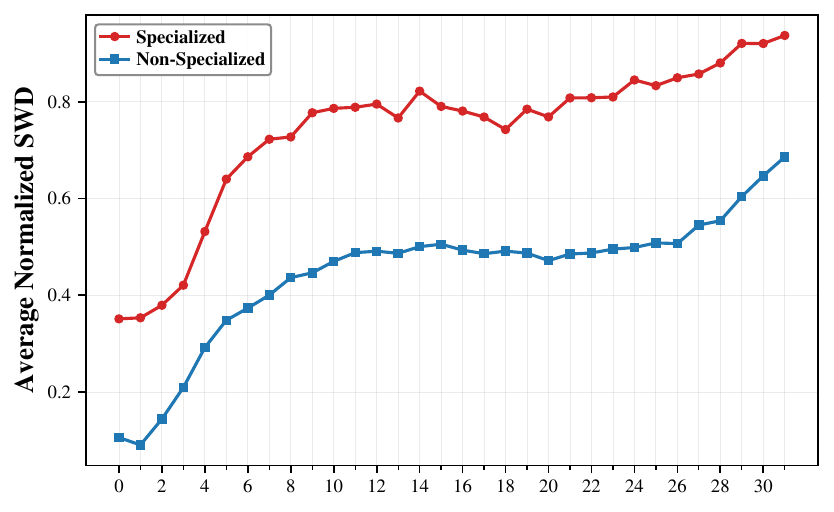}
        \vspace{-0.24in}
        \captionof{figure}{Layer-wise embedding distribution distance under specialized and non-specialized tokens using sliced Wasserstein distances (SWD)~\citep{swd}.}
        \label{fig:specialized_vs_non_specialized}
    \end{minipage}
\end{figure*}

To examine whether embedding-space signals can indeed reveal modality specialization, we analyze how modality-specific inputs are processed by the base LLM and modality-specialized MLLMs. As shown in Fig.~\ref{fig:mol_embedding_visualization}, when molecular tokens are processed by different MLLMs, the hidden representations form clearly different embedding distributions. In particular, the molecule LLM induces a more distinct distribution due to its modality-specific understanding. 
Further, we measure the embedding distribution distance between the base LLM and each specialized MLLM under specialized and non-specialized token inputs. Fig.~\ref{fig:specialized_vs_non_specialized} shows that specialized inputs consistently produce larger embedding distribution distance than non-specialized inputs, suggesting that the embedding space faithfully reflects modality specialization. Based on this analysis, we use embedding-space discrepancies induced by modality-token inputs as input-conditioned signals for estimating merging coefficients.

Motivated by this observation, we propose an embedding-signal-based MLLM merging (\textbf{ES-Merging}), a novel framework that moves the model merging paradigm from parameter space signals to embedding space signals. Our intuition is that the modality specialization of an MLLM can be measured by the differences in the embedding space between the base LLM and the MLLM as shown in Fig.~\ref{fig:mol_embedding_visualization}. To this end, we first design a probe input containing multimodal tokens of different modalities and obtain layer-wise embeddings that reflect modality-specific representation changes across layers. Based on the embedding differences between each MLLM and base LLM, we estimate merging coefficients at two complementary granularity levels: layer-wise and element-wise. The layer-wise coefficients capture coarse-grained modality specialization across transformer layers, whereas the element-wise coefficients capture fine-grained specialization within each layer. By integrating these complementary signals, ES-Merging produces calibrated merging coefficients that determine how much each modality-specialized MLLM contributes to the unified MLLM.

We validate ES-Merging by merging three modality-specialized biological MLLMs, covering molecule, protein, and cell modalities, into a single unified model. Through extensive experiments, we demonstrate that ES-Merging substantially outperforms parameter-signal-based merging methods on cross-modal reasoning as well as on single-modal knowledge preservation. Further analyses show that ES-Merging is an efficient counterpart to task-specific fine-tuning by achieving a 5.95$\times$ reduction in computational cost, while providing explainability along with stronger or comparable performance. Additional experiments on image, video, and audio modalities suggest that embedding space signals can provide useful guidance beyond the biological domain. These results highlight that the embedding space provides principled and efficient signals for merging MLLMs.
\section{Related Work}
\paragraph{MLLMs for Science}
Large language models (LLMs)~\citep{touvron2023llama2,llama3,openai2024gpt4,openai2024gpt4ocard,comanici2025gemini} are increasingly being extended to scientific problems through multimodal large language models (MLLMs) that incorporate diverse biological modalities, including molecules~\citep{moleculesurvey}, proteins~\citep{proteinsurvey}, and cells~\citep{cellsurvey}. In the protein domain, prior works have modeled amino-acid sequences~\citep{protst,biot5} or jointly with structures~\citep{proteinchat,prot2text,protchatgpt,proteingpt,prot2chat,prot2textv2}. Single cell LLMs learn the scRNA-seq representations~\citep{instructcell,cell2text,cello1} or further incorporate histology information~\citep{spallm,spatial2sentence}. On the other hand, molecular LLMs have been built on 1D string representations~\citep{chemberta,molt5} such as SMILES~\citep{smiles} and SELFIES~\citep{selfies}, 2D molecular graphs~\citep{molca,molinstructions,instructmol,llasmol,llamo}, 3D structures with conformations~\citep{3dmolm,unimot}, and joint 2D-3D representations with multiple encoders~\citep{molllama}. Despite their progress, existing biological MLLMs are limited to understand and process a single modality, hindering their ability to solve inherently cross-modal scientific problems.

\paragraph{Model Merging}
Model merging aims to integrate multiple specialist models into a unified model with minimal additional training. Existing methods typically guide merging through parameter-space signals, including magnitude or agreement of parameter updates~\citep{taskarithmetic,consensusmerging,pcbmerging}, sign alignment or masking for resolving update conflicts~\citep{tiesmerging,emrmerging}, learned layer-wise coefficients~\citep{adamerging}, input-dependent routing~\citep{twinmerging}, and latent-space interpolation~\citep{lsmerge}. Recent MLLM merging methods also remain largely grounded in parameter- or weight-space operations, such as improving the robustness of low-rank module merging~\citep{robustmerge} or using uncertainty to select and order vision-language MLLMs before weight-space merging~\citep{uqmerge}. However, these methods do not directly estimate modality specialization from embedding space responses induced by modality-specific token inputs. We instead propose a merging framework that derives layer-wise and element-wise merging coefficients from embedding signals, enabling modality-aware merging of biological MLLMs specialized to heterogeneous modalities.

\section{Preliminary}
We begin by formally describing MLLMs, then formulating the model merging with LoRA.

\paragraph{Multimodal Large Language Model\label{sec:MLLM}}
MLLMs operate beyond the textual space by taking modality token embeddings with text tokens. Concretely, let $\mathcal{M} = \{m_1, ..., m_K\}$ denote a set of modalities. For each modality $m_i$, a modality-specific encoder $f_{m_i}$ projects the raw modality input $\mathbf{x}_{m_i}$ to a sequence of modality tokens: $\mathbf{H}_{m_i} = f_{m_i}(\mathbf{x}_{m_i})$. Then, an MLLM generates textual output $\mathbf{y}$ by taking the concatenated textual and modality tokens: $\mathbf{y} = g(\mathbf{H}^0), \; \text{where} \; \mathbf{H}^0 = [\mathbf{H}_{\text{text}}; \mathbf{H}_{m_i}; ...].$
In this view, the core interface of an MLLM is the token-based embeddings: raw modality inputs are first mapped into sequences of vectors that are compatible with the LLM's embedding dimension, and then transformed over layers to elicit modality-specific behaviors.

\paragraph{LoRA Merging}
In the MLLM setting, modality specialization is often implemented with LoRA~\citep{hu2022lora}, enabling unified model construction via weighted LoRA merging. Let $\theta_{m_i}$ denote the LoRA parameters specialized to modality $m_i$.
We index the parameters by layer and weight element: $\theta^{l,n}_{m_i}$ denotes the $n$-th weight in the $l$-th layer of $\theta_{m_i}$.
We obtain the unified LoRA parameters as a weighted summation: $\theta^{l,n}_{\mathrm{uni}} \leftarrow \sum_{m_i \in \mathcal{M}} \lambda^{l,n}_{m_i}\theta^{l,n}_{m_i}$~\cite{lorahub}.
Under this formulation, the main challenge is to estimate appropriate merging coefficients $\lambda^{l,n}_{m_i}$ that determine how strongly each modality-specific parameter contributes to modality understanding. Our method addresses this coefficient estimation problem by deriving $\lambda^{l,n}_{m_i}$ from embedding space signals rather than parameter space statistics.

\section{ES-Merging}
We present Embedding-Signal-based MLLM Merging (ES-Merging), a representation-aware framework for merging modality-specialized MLLMs based on embedding space signals. Specifically, ES-Merging constructs probe inputs to elicit representational differences across models, and converts these embedding space discrepancies into layer-wise and element-wise merging coefficients.

\subsection{Probe Input}
Probe inputs are designed to provide a common diagnostic context in which the base LLM and all modality-specialized MLLMs process the same set of modality-token embeddings. Rather than using downstream task prompts, which may introduce task-specific biases, we construct probe inputs only to elicit the representation transformations induced by modality tokens. This allows us to measure modality specialization from the backbone response itself, while keeping the input context shared across different specialized models.
Specifically, for each modality $m_i$, we sample raw inputs $\mathbf{x}_{m_i}$ from the corresponding dataset and obtain modality token embeddings as $\mathbf{H}_{m_i}=f_{m_i}(\mathbf{x}_{m_i})$. We construct each probe input by concatenating short textual prefixes with modality token blocks, i.e., $\mathbf{H}_{\text{probe}}^{0}=[\mathbf{H}_{\text{text},m_1};\mathbf{H}_{m_1};\mathbf{H}_{\text{text},m_2};\mathbf{H}_{m_2};\ldots]$, where $\mathbf{H}_{\text{text},m_i}$ denotes the text-prefix embeddings.

\begin{wrapfigure}{r}{0.37\textwidth}
\centering
\begin{tikzpicture}
\node[draw, dashed, rounded corners=3pt, inner sep=7pt, text width=0.33\textwidth] (content) {
\texttt{\small Molecule: \textcolor{red}{<molecule\_tokens>}}\\
\texttt{\small Protein: \textcolor{blue}{<protein\_tokens>}}\\
\texttt{\small Cell: \textcolor{green!50!black}{<cell\_tokens>}}
};
\end{tikzpicture}
\vspace{-0.15in}
\caption{Probe input template.}
\vspace{-0.15in}
\label{fig:prompt-template}
\end{wrapfigure}

In the biological setting, the modality set $\mathcal{M}$ typically consists of $\{\textcolor{red}{\mathrm{molecule}}, \textcolor{blue}{\mathrm{protein}}, \textcolor{green!50!black}{\mathrm{cell}}\}$, and the resulting probe input is illustrated in Fig.~\ref{fig:prompt-template}. By forwarding each probe input through the base LLM and each MLLM specialized to modality $m_j$, we obtain layer-wise embeddings by extracting the modality $m_i$ tokens, denoted as $\mathbf{H}_{m_i \rightarrow \text{base}}^{l}$ and $\mathbf{H}_{m_i \rightarrow \theta_{m_j}}^{l}$.

These layer-wise representations serve as the basis for estimating merging coefficients. We regard the embedding discrepancy between the base LLM and each modality-specialized MLLM as an embedding space signal that reflects the degree of modality specialization. ES-Merging uses this signal at two complementary levels: globally across layers to estimate layer-wise specialization, and locally within each layer to identify parameter elements associated with specialized transformations.

\begin{figure*}[t]
    \centering
    \includegraphics[width=\linewidth]{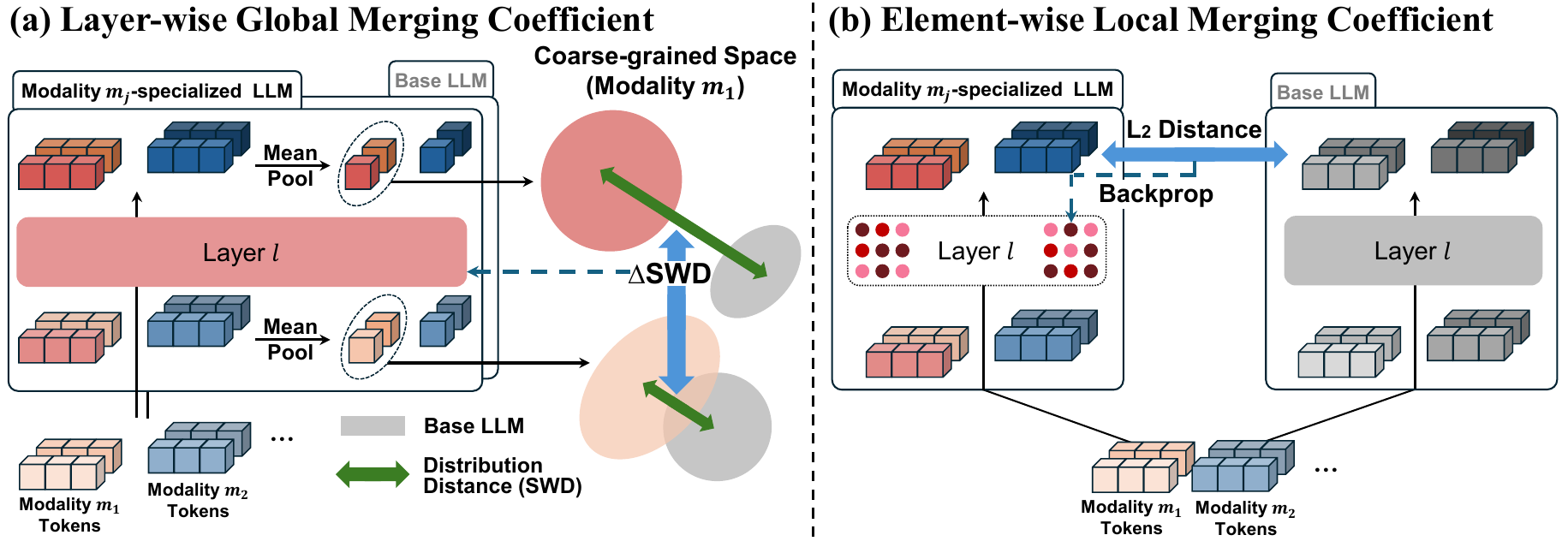}
    \vspace{-0.2in}
    \caption{Overview of ES-Merging. (A) Layer-wise global merging coefficients are computed from the coarse-grained embedding signals, which are the layer-wise differences of distribution distances between mean pooled representations of the base LLM and a specialized MLLM. (B) Element-wise local merging coefficients are assigned based on the fine-grained embedding signals by computing the gradients from the embedding-wise distances.}
    \label{fig:method}
\end{figure*}

\subsection{Layer-wise Global Merging Coefficient}
\label{sec:layerwise-global}

We first estimate the layer-wise contribution of each modality-specialized MLLM to the merged model from coarse-grained embedding signals (Fig.~\ref{fig:method} (a)). We aim to assign a larger merging coefficient when a layer induces stronger representational transformations from the base LLM across modality inputs, reflecting the substantial modality-specific transformations.

\paragraph{Coarse-grained Embedding Signal}
Given the layer-wise modality-token embeddings from the probe inputs, $\mathbf{H}_{m_i \rightarrow \text{base}}^{l}$ and $\mathbf{H}_{m_i \rightarrow \theta_{m_j}}^{l}$, we first apply mean pooling over the token dimension to summarize each representation. This gives a coarse-grained representation of how the base LLM and each modality-specialized MLLM process the same modality tokens at each layer. We then measure the distributional gap between the representations of the base LLM and specialized MLLMs using sliced Wasserstein distance (SWD)~\citep{swd}. Since the SWD at layer $l$ reflects the cumulative representational shift up to that layer, we use the layer-wise differences, $\Delta\mathrm{SWD}^{l}_{m_i\rightarrow\theta_{m_j}}=\mathrm{SWD}^{l}_{m_i\rightarrow\theta_{m_j}}-\mathrm{SWD}^{l-1}_{m_i\rightarrow\theta_{m_j}}$, as the coarse-grained embedding signal. A larger value indicates that the corresponding layer introduces a stronger modality-specific transformation.

\paragraph{Layer-wise Coefficient Estimation}
To convert the coarse-grained signals into merging coefficients, we normalize the layer-wise SWD differences and then aggregate them over input modalities to obtain the layer-wise importance score of each specialized model, defined as $s^l_{\theta_{m_j}}=\sum_{m_i\in\mathcal{M}}\widehat{\Delta\mathrm{SWD}}^{l}_{m_i\rightarrow\theta_{m_j}}$. This score reflects how strongly $\theta_{m_j}$ contributes to coarse-grained representational transformations over different modality inputs at layer $l$. We then apply a softmax across specialized models to obtain the layer-wise global merging coefficient: $\alpha_{m_j}^l=\frac{\exp(s^l_{\theta_{m_j}} / \tau)}{\sum_{m \in \mathcal{M}}\exp(s^l_{\theta_{m}} / \tau)}.$
Thus, each specialized MLLM receives a larger layer-wise coefficient at layers where it induces stronger coarse-grained representational changes.

\subsection{Element-wise Local Merging Coefficient}
\label{sec:elementwise-local}

While the layer-wise global coefficients capture the contribution of each specialized model at the transformer-layer level, it assigns the merging coefficient uniformly to all parameters within a layer. To capture fine-grained contributions of each parameter, we further introduce an element-wise local merging coefficient from fine-grained embedding signals (Fig.~\ref{fig:method} (b)).

\paragraph{Fine-grained Embedding Signal}
Unlike the coarse-grained signal, which is computed from mean-pooled representations, the fine-grained signal directly measures discrepancies between modality-token representations. For each modality $m_i$ and specialized model $\theta_{m_j}$, we compute the Frobenius distance between $\mathbf{H}_{m_i \rightarrow \text{base}}^{l}$ and $\mathbf{H}_{m_i \rightarrow \theta_{m_j}}^{l}$, averaged over the probe inputs. We denote this fine-grained discrepancy as $r^{l}_{m_i \rightarrow \theta_{m_j}}$. This signal reflects how differently the specialized model processes modality-$m_i$ tokens relative to the base LLM at layer $l$.

\paragraph{Element-wise Coefficient Estimation}
To derive the parameter-level merging coefficients, we identify which parameter most affect to the fine-grained embedding discrepencies between base LLM and the specialized MLLMs, $r^l_{m_i \rightarrow \theta_{m_j}}$. Specifically, we regard $r^l_{m_i \rightarrow \theta_{m_j}}$ as a local alignment signal and, by computing the gradient of this discrepancy with respect to each parameter element, we obtain the element-wise importance score, $s^{l,n}_{\theta_{m_j}}=\sum_{m_i \in \mathcal{M}}\left|\partial r^{l}_{m_i \rightarrow \theta_{m_j}}/\partial \theta_{m_j}^{l,n}\right|$. A larger score indicates that the corresponding element has a stronger local influence on the base-specialized embedding gap, and thus is more closely tied to modality-specialized representation changes. We normalize theses scores and then apply a softmax across specialized models to obtain the element-wise local merging coefficient: $\beta^{l,n}_{m_j}=\frac{\exp(\hat{s}^{l,n}_{\theta_{m_j}}/\tau)}{\sum_{m \in \mathcal{M}}\exp(\hat{s}^{l,n}_{\theta_m}/\tau)}.$ Each parameter element receives a larger coefficient when it more strongly governs fine-grained embedding-space discrepancies.

\subsection{Integrating Layer- and Element-wise Merging Coefficients}
\label{sec:mixing-coefficients}

The coarse-grained layer-wise coefficient $\alpha^{l}_{m_i}$ and the fine-grained element-wise coefficient $\beta^{l,n}_{m_i}$ each capture modality-specific importance at different levels of granularity. To complementarily combine two different coefficients, we compute a final merging coefficient by multiplying the layer-wise coefficient and the element-wise coefficient and renormalizing across modalities as follows:
\begin{align*}
\lambda^{l,n}_{m_i} = \frac{\alpha^{l}_{m_i} \cdot \beta^{l,n}_{m_i}}{\sum_{m \in \mathcal{M}} \alpha^{l}_{m} \cdot \beta^{l,n}_{m}}.
\label{eq:gamma_comprehensive}
\end{align*}

By integrating different coefficient types from different granularities of embedding signals, ES-Merging enables calibrated merging coefficients that preserve complementary modality expertise more faithfully, enabling robust cross-modal knowledge composition. 
\newcommand{\avgcell}[1]{\cellcolor{gray!15}#1}

\begin{table*}[t]
\centering
\caption{Performance comparison of ES-Merging with the base, specialized MLLMs, and merging methods on the instance-varying interaction prediction tasks. We report accuracy and macro-F1 across each subset. \textbf{Bold} and \underline{underline} indicate the best and second best performances, respectively.}
\vspace{-0.05in}
\renewcommand{\arraystretch}{1.15}
\renewcommand{\tabcolsep}{4pt}
\resizebox{\textwidth}{!}{%
\scriptsize
\begin{tabular}{l *{14}{c}}
\toprule
& \multicolumn{8}{c}{\textbf{Molecule-Protein Interaction}} 
& \multicolumn{6}{c}{\textbf{Molecule-Cell Interaction}} \\
\cmidrule(l{2pt}r{2pt}){2-9} 
\cmidrule(l{2pt}r{2pt}){10-15}
& \multicolumn{2}{c}{BindingDB} 
& \multicolumn{2}{c}{BioSNAP} 
& \multicolumn{2}{c}{Human} 
& \multicolumn{2}{>{\columncolor{gray!15}}c}{\textit{Avg.}} 
& \multicolumn{2}{c}{DrugComb} 
& \multicolumn{2}{c}{GDSC2} 
& \multicolumn{2}{>{\columncolor{gray!15}}c}{\textit{Avg.}} \\
\cmidrule(l{2pt}r{2pt}){2-3} 
\cmidrule(l{2pt}r{2pt}){4-5} 
\cmidrule(l{2pt}r{2pt}){6-7} 
\cmidrule(l{2pt}r{2pt}){8-9} 
\cmidrule(l{2pt}r{2pt}){10-11} 
\cmidrule(l{2pt}r{2pt}){12-13} 
\cmidrule(l{2pt}r{2pt}){14-15}
& Acc. & F1 
& Acc. & F1 
& Acc. & F1 
& \avgcell{Acc.} & \avgcell{F1} 
& Acc. & F1 
& Acc. & F1 
& \avgcell{Acc.} & \avgcell{F1} \\
\midrule

\rowcolor{lightgreen}
\multicolumn{15}{l}{\textit{Base LLM and Specialized MLLMs}} \\

LLaMA-3.1-8B-Instruct  
& 51.9 & 51.7 
& 61.5 & 60.2 
& 59.0 & 58.8 
& \avgcell{57.5} & \avgcell{56.9} 
& 73.7 & 73.6 
& 84.8 & 84.7 
& \avgcell{79.3} & \avgcell{79.1} \\

Mol-LLaMA~\citep{molllama}              
& 55.8 & 54.7 
& 66.4 & 64.0 
& \underline{61.5} & \underline{61.4} 
& \avgcell{61.2} & \avgcell{60.0} 
& 64.1 & 64.1 
& 55.5 & 55.5 
& \avgcell{59.8} & \avgcell{59.8} \\

Prot2Text-V2~\citep{prot2textv2}           
& 59.2 & 59.2 
& 55.3 & 54.1 
& 47.2 & 47.2 
& \avgcell{53.9} & \avgcell{53.5} 
& 65.6 & 65.6 
& 59.7 & 57.9 
& \avgcell{62.6} & \avgcell{61.7} \\

Cell-o1~\citep{cello1}                
& 53.5 & 51.5 
& 59.3 & 59.3 
& 49.1 & 48.0 
& \avgcell{54.0} & \avgcell{52.9} 
& 76.3 & 76.2 
& 85.9 & 85.8 
& \avgcell{81.1} & \avgcell{81.0} \\

\midrule

\rowcolor{lightgreen}
\multicolumn{15}{l}{\textit{Merging Methods}} \\

Avg. Merging                         
& \underline{65.3} & \underline{64.9} 
& \underline{67.4} & 66.5 
& 60.9 & 60.9 
& \avgcell{\underline{64.5}} & \avgcell{\underline{64.1}} 
& 72.5 & 72.4 
& 85.4 & 85.2 
& \avgcell{78.9} & \avgcell{78.8} \\

TIES-Merging~\citep{tiesmerging}      
& 60.8 & 60.7 
& 62.7 & 61.6 
& 58.6 & 58.5 
& \avgcell{60.7} & \avgcell{60.3} 
& 74.7 & 74.6 
& 85.9 & 85.7 
& \avgcell{80.3} & \avgcell{80.2} \\

EMR-Merging~\citep{emrmerging}        
& 64.7 & 64.2 
& 67.3 & \underline{66.8} 
& 60.4 & 60.4 
& \avgcell{64.1} & \avgcell{63.8} 
& 45.2 & 45.0 
& 93.4 & 93.3 
& \avgcell{69.3} & \avgcell{69.2} \\

AdaMerging~\citep{adamerging}         
& 52.2 & 51.0 
& 64.2 & 61.7 
& 59.9 & 59.8 
& \avgcell{58.8} & \avgcell{57.5} 
& 16.0 & 14.1 
& 17.2 & 15.3 
& \avgcell{16.6} & \avgcell{14.7} \\

PCB-Merging~\citep{pcbmerging}        
& 55.1 & 55.1 
& 60.3 & 58.8 
& 58.6 & 58.4 
& \avgcell{58.0} & \avgcell{57.4} 
& 77.3 & 77.3 
& 86.0 & 85.8 
& \avgcell{81.7} & \avgcell{81.6} \\

Consensus Merging~\citep{consensusmerging}
& 59.2 & 59.2 
& 62.1 & 60.9 
& 59.2 & 59.1 
& \avgcell{60.2} & \avgcell{59.7} 
& 76.0 & 76.0 
& 84.3 & 84.2 
& \avgcell{80.2} & \avgcell{80.1} \\

UQ-Merge~\citep{uqmerge}              
& 43.7 & 40.7
& 56.5 & 51.1
& 49.3 & 48.4
& \avgcell{49.8} & \avgcell{46.8}
& 14.0 & 13.8
& 71.7 & 66.5
& \avgcell{42.8} & \avgcell{40.1} \\

RobustMerge~\citep{robustmerge}       
& 42.8 & 42.8
& 49.0 & 48.2
& 43.9 & 43.4
& \avgcell{45.2} & \avgcell{44.8}
& 11.0 & 10.8
& 9.5 & 9.4
& \avgcell{10.3} & \avgcell{10.1} \\

LS-Merge~\citep{lsmerge}              
& 52.3 & 52.1 
& 61.8 & 60.6 
& 58.8 & 58.5 
& \avgcell{57.6} & \avgcell{57.1} 
& 77.3 & 77.3 
& 84.5 & 84.4 
& \avgcell{80.9} & \avgcell{80.8} \\

\cdashline{1-15}[0.5pt/1pt]\rule{-2pt}{2.5ex}

Avg. Merging + FT 
& 60.5 & 59.5 
& 55.8 & 55.8 
& 57.2 & 57.2 
& \avgcell{57.8} & \avgcell{57.5} 
& \textbf{81.1} & \textbf{80.8} 
& \underline{94.0} & \underline{93.9} 
& \avgcell{\textbf{87.5}} & \avgcell{\textbf{87.4}} \\

\cdashline{1-15}[0.5pt/1pt]\rule{-2pt}{2.5ex}

\textbf{ES-Merging (Ours)} 
& \textbf{66.0} & \textbf{65.3} 
& \textbf{69.1} & \textbf{68.4} 
& \textbf{62.0} & \textbf{61.9} 
& \avgcell{\textbf{65.7}} & \avgcell{\textbf{65.2}} 
& \underline{80.7} & \underline{80.2} 
& \textbf{94.1} & \textbf{94.0} 
& \avgcell{\underline{87.4}} & \avgcell{\underline{87.1}} \\

\bottomrule
\end{tabular}%
}
\label{tab:general_interaction_prediction}
\vspace{-0.05in}
\end{table*}
\section{Experimental Results}

\subsection{Experimental Setup\label{sec:experimental_setup}}

\paragraph{Implementation Details}
For merging, we leverage the state-of-the-art MLLMs specialized to each modality: Mol-LLaMA~\citep{molllama} for molecule modality, Prot2Text-V2~\citep{prot2textv2} for protein modality, and Cell-o1~\citep{cello1} for single cell modality, whose base LLMs are LLaMA-3.1-8B-Instruct~\citep{llama3}. The merging methods are applied across all modality-specific models, resulting in a unified MLLM. Since each downstream task requires distinct domain knowledge, we provide task-specific few-shot in-context examples to induce the appropriate task understanding. For a fair comparison, we use the same instruction templates and in-context examples for all compared methods and leverage the greedy decoding during the LLM inference to reduce randomness. For more details about the implementations and the prompts, please refer to Appendix~\ref{sec:impl_details} and \ref{sec:prompt_setting}, respectively.

\paragraph{Baselines}
We compare ES-Merging with the base LLM, MLLMs specialized to a single modality, and merging methods. As modality-specialized baselines, we consider Mol-LLaMA, Prot2Text-V2, and Cell-o1, each specialized for molecule, protein, and single-cell modalities, respectively. We also compare ES-Merging with parameter-signal-based merging methods, including Average Merging, TIES-Merging~\citep{tiesmerging}, EMR-Merging~\citep{emrmerging}, layer-wise AdaMerging~\citep{adamerging}, PCB-Merging~\citep{pcbmerging}, Consensus Merging~\citep{consensusmerging}, LS-Merge~\citep{lsmerge}, and task-specific finetuned model after the average merging, denoted as Avg. Merging + FT. For the base LLMs and the specialized MLLMs, unsupported modalities are represented as textual inputs. Baseline details are provided in Appendix~\ref{sec:base_details}.

\subsection{Instance-varying Interaction Prediction~\label{sec:instance_varying_interaction_prediction}}

We first evaluate on instance-varying cross-modal interaction tasks, where the target counterpart changes across instances and the model should generalize across diverse cross-modal pairs.

\paragraph{Datasets}
We consider two instance-varying cross-modal interaction settings: molecule-protein and molecule-cell interactions. For molecule-protein interaction, the task is to predict whether a given molecule interacts with a given protein, including BindingDB, BioSNAP, and Human~\citep{psichic}. For molecule-cell interaction, the task is to predict the effect of molecules on a given cell, including DrugComb~\citep{drugcomb} and GDSC2~\citep{gdsc2}. In both settings, the molecule, protein, and cell counterpart changes for each instance, requiring the model to generalize across diverse cross-modal combinations rather than relying on a fixed target. We provide the detailed explanation of datasets in Appendix~\ref{sec:data_details}.

\paragraph{Results}
As shown in Table~\ref{tab:general_interaction_prediction}, ES-Merging consistently outperforms parameter-signal-based merging baselines, suggesting that ES-Merging effectively integrates complementary cross-modal knowledge from modality-specialized MLLMs and thus leading to stronger generalization when the interaction counterpart varies across instances. Interestingly, ES-Merging outperforms the task-specific finetuned model (Avg. Merging + FT) on the molecule-protein interaction tasks and shows comparable performance on the molecule-cell interaction tasks, indicating that ES-merging can enhance cross-modal reasoning without further downstream finetuning. On the other hand, existing merging baselines such as EMR-Merging or RobustMerge often exhibit instability across datasets. These results imply that input-blind parameter space heuristics are insufficient for reliably combining heterogeneous multimodal specialization, supporting that ES-Merging provides a more robust integration of modality-specific knowledge and behaviors from the input-aware embedding space signals.

\begin{table*}
\centering
\caption{Performance comparison of ES-Merging with the base, specialized MLLMs, and merging methods on the target-fixed functionality prediction tasks. We report accuracy and macro-F1 across each subset. \textbf{Bold} indicates the best and \underline{underline} indicates the second best.}
\vspace{-0.05in}
\renewcommand{\arraystretch}{1.15}
\renewcommand{\tabcolsep}{4pt}
\resizebox{\textwidth}{!}{%
\scriptsize
\begin{tabular}{l c c c c c c c c c c a a c c c c c c a a}
\toprule
& \multicolumn{12}{c}{\textbf{CYP Inhibition}} & \multicolumn{8}{c}{\textbf{CYP Substrate}} \\
\cmidrule(l{2pt}r{2pt}){2-13} \cmidrule(l{2pt}r{2pt}){14-21}
& \multicolumn{2}{c}{CYP1A2} & \multicolumn{2}{c}{CYP2C19} & \multicolumn{2}{c}{CYP2C9} & \multicolumn{2}{c}{CYP2D6} & \multicolumn{2}{c}{CYP3A4} & \multicolumn{2}{c}{\textit{Avg.}} & \multicolumn{2}{c}{CYP2C9} & \multicolumn{2}{c}{CYP2D6} & \multicolumn{2}{c}{CYP3A4} & \multicolumn{2}{c}{\textit{Avg.}} \\
\cmidrule(l{2pt}r{2pt}){2-3} \cmidrule(l{2pt}r{2pt}){4-5} \cmidrule(l{2pt}r{2pt}){6-7} \cmidrule(l{2pt}r{2pt}){8-9} \cmidrule(l{2pt}r{2pt}){10-11} \cmidrule(l{2pt}r{2pt}){12-13} \cmidrule(l{2pt}r{2pt}){14-15} \cmidrule(l{2pt}r{2pt}){16-17} \cmidrule(l{2pt}r{2pt}){18-19} \cmidrule(l{2pt}r{2pt}){20-21}
 & Acc. & F1 & Acc. & F1 & Acc. & F1 & Acc. & F1 & Acc. & F1 & Acc. & F1 & Acc. & F1 & Acc. & F1 & Acc. & F1 & Acc. & F1 \\

\midrule
\rowcolor{lightgreen}
\multicolumn{21}{l}{\textit{Base LLM and Specialized MLLMs}} \\
LLaMA-3.1-8B-Instruct  & 58.9 & 55.3 & 56.0 & 50.6 & 55.9 & 47.4 & 61.2 & 56.4 & 55.1 & 43.1 & 57.4 & 50.6 & 57.7 & 47.8 & 34.2 & 30.1 & 55.6 & 48.5 & 49.2 & 42.1 \\
Mol-LLaMA              & 68.6 & 67.1 & 65.8 & 64.3 & 63.3 & 63.3 & 64.7 & 59.2 & 64.3 & 64.0 & 65.3 & 63.6 & 62.7 & 49.7 & \textbf{61.7} & \textbf{57.4} & 55.2 & 53.7 & 59.9 & \underline{53.6} \\
Prot2Text-V2           & 59.4 & 55.8 & 56.8 & 52.5 & \underline{67.4} & 60.1 & \underline{76.3} & 55.9 & 59.6 & 50.3 & 63.9 & 54.9 & 54.5 & 47.5 & 53.4 & 51.4 & 56.0 & 53.8 & 54.6 & 50.9 \\
Cell-o1                & 53.5 & 53.5 & 44.9 & 44.9 & 43.2 & 40.6 & 46.9 & 40.9 & 43.6 & 43.5 & 46.4 & 44.7 & 36.6 & 34.3 & 36.1 & 34.9 & 48.5 & 47.5 & 40.4 & 38.9 \\

\midrule
\rowcolor{lightgreen}
\multicolumn{21}{l}{\textit{Merging Methods}} \\
Avg. Merging            & 61.2 & 60.4 & 54.1 & 53.2 & 52.7 & 52.5 & 53.8 & 49.5 & 54.6 & 54.6 & 55.3 & 54.0 & 40.3 & 38.6 & 42.1 & 41.3 & 49.3 & 45.8 & 43.9 & 41.9 \\
TIES-Merging            & \underline{71.8} & \underline{71.7} & \underline{67.8} & \underline{67.7} & 66.5 & 63.5 & 76.1 & 63.5 & 65.3 & 64.8 & \underline{69.5} & 66.2 & 56.0 & 43.5 & 55.6 & 52.1 & 54.5 & 53.1 & 55.4 & 49.6 \\
EMR-Merging             & 70.4 & 70.4 & 64.8 & 64.7 & 66.8 & \underline{64.5} & 74.2 & 61.7 & 66.1 & 65.6 & 68.5 & 65.4 & 60.5 & 49.2 & 55.6 & 52.9 & 51.5 & 49.4 & 55.9 & 50.5 \\
AdaMerging              & 52.5 & 49.4 & 53.1 & 50.3 & 44.8 & 43.6 & 45.6 & 43.9 & 50.5 & 49.4 & 49.3 & 47.3 & 33.6 & 33.6 & 43.6 & 43.4 & 49.3 & 45.3 & 42.2 & 40.8 \\
PCB-Merging             & 69.2 & 69.0 & 63.6 & 63.3 & 65.0 & 61.8 & 72.4 & 60.1 & 63.8 & 63.0 & 66.8 & 63.4 & 55.2 & 44.9 & 57.1 & 53.3 & \underline{56.7} & 55.0 & 56.4 & 51.1 \\
Consensus Merging       & 68.5 & 68.2 & 63.6 & 63.3 & 65.7 & 62.3 & 73.5 & 60.2 & 63.8 & 62.9 & 67.0 & 63.4 & 54.5 & 45.3 & 55.6 & 52.9 & \underline{56.7} & \underline{55.7} & 55.6 & 51.3 \\
UQ-Merge                & 50.3 & 49.5 & 48.1 & 36.4 & 36.2 & 29.8 & 26.1 & 24.9 & 42.7 & 32.6 & 40.5 & 34.6 & 26.1 & 24.9 & 28.6 & 26.9 & 49.3 & 49.3 & 34.7 & 33.7 \\
RobustMerge             & 41.2 & 40.6 & 39.6 & 39.4 & 36.6 & 36.6 & 39.2 & 35.9 & 40.8 & 40.8 & 39.5 & 38.7 & 50.0 & 46.8 & 45.9 & 45.3 & 44.8 & 41.9 & 46.9 & 44.7 \\
LS-Merge                & 55.3 & 55.3 & 51.0 & 51.0 & 47.6 & 46.3 & 56.1 & 48.5 & 44.1 & 44.1 & 50.8 & 49.0 & 44.0 & 40.9 & 39.1 & 38.4 & 41.8 & 39.8 & 41.6 & 39.8 \\
\cdashline{1-21}[0.5pt/1pt]\rule{-2pt}{2.5ex}
Avg. Merging + FT       & 68.3 & 68.3 & 67.0 & 67.0 & 62.3 & 62.2 & 67.6 & \underline{67.6} & \underline{67.7} & \underline{67.6} & 66.6 & \underline{66.5} & \textbf{65.7} & \underline{50.6} & 59.4 & 53.0 & 55.2 & 54.6 & \underline{60.1} & 52.7 \\
\cdashline{1-21}[0.5pt/1pt]\rule{-2pt}{2.5ex}
\textbf{ES-Merging (Ours)} & \textbf{77.4} & \textbf{77.4} & \textbf{70.6} & \textbf{70.5} & \textbf{72.5} & \textbf{70.4} & \textbf{80.7} & \textbf{69.5} & \textbf{71.3} & \textbf{70.8} & \textbf{74.5} & \textbf{71.7} & \underline{64.2} & \textbf{53.6} & \underline{60.9} & \underline{57.2} & \textbf{60.5} & \textbf{59.6} & \textbf{61.9} & \textbf{56.8} \\
\bottomrule
\end{tabular}%
}
\label{tab:cyp_interaction_prediction}
\end{table*}

\subsection{Target-fixed Functionality Prediction~\label{sec:target_fixed_interaction_prediction}}
We further evaluate on target-fixed cross-modal functionality prediction tasks, where each subtask is associated with a fixed target and requires target-specific biological functionality knowledge beyond simple interaction matching in Section~\ref{sec:instance_varying_interaction_prediction}.

\paragraph{Datasets}
To this end, we evaluate on CYP enzyme prediction. Unlike the instance-varying setting above, each subtask is defined with respect to a fixed enzyme target and the tasks are predicting the biological functionality of the given molecule to the fixed target: inhibitory effects or substrate specificity. Specifically, we consider five CYP inhibition subtasks to CYP1A2, CYP2C19, CYP2C9, CYP2D6, and CYP3A4 enzymes~\citep{cypinhibit}, and three CYP substrate subtasks for CYP2C9, CYP2D6, and CYP3A4 enzymes~\citep{cypstrate}, from the TDC dataset~\citep{tdc}. These benchmarks therefore evaluate whether the merged model can capture not only detailed molecular structural variation under a shared target, but also biologically meaningful interaction types.

\paragraph{Results}
As shown in Table~\ref{tab:cyp_interaction_prediction}, ES-Merging achieves the best average performance, suggesting that it more effectively preserves and integrates expert knowledge required for target-specific functionality prediction. In contrast, on the CYP substrate tasks, most merging baselines based on the parameter signals underperform Mol-LLaMA, because these tasks largely rely on molecular structural understanding under a fixed-target setting. ES-Merging achieves performance comparable to or better than Mol-LLaMA, indicating that embedding-signal-based merging preserves molecular structural understanding while incorporating target-specific functional knowledge from expert models.

\subsection{Single-modal Knowledge Preservation}
To see whether ES-Merging can preserve the knowledge that is learned by each specialized MLLMs, we evaluate merging methods on single-modal tasks beyond cross-modal tasks. We leverage PAMPA~\cite{pampa}, Immuno-Virus~\cite{venusvaccine}, and PBMC~\cite{pbmc} as single-modal tasks related to molecules, proteins, and cells, respectively. For detailed experimental setup, please refer to Appendix~\ref{sec:prompt_setting}.

\paragraph{Results} As shown in Table~\ref{tab:single_modality_binary_prediction}, ES-Merging substantially improves single-modal knowledge preservation over existing merging methods, achieving the highest average accuracy among all merged models, indicating that embedding space signals help mitigate the degradation of modality-specific knowledge that commonly occurs when heterogeneous experts are merged in parameter space. Notably, ES-Merging marginally surpasses the corresponding specialized MLLMs on the molecule and protein tasks. These results suggest that ES-Merging not only preserves expert knowledge, but also constructs a more effective shared representation by integrating complementary information from other modality-specialized models. On the cell task, ES-Merging achieves the best performance among all merging methods, although it remains below the cell-specialized MLLM. This is mainly due to the little transferability of non-cell MLLMs to cell-related tasks as discussed in Appendix~\ref{app:sec:single_modal_detailed_analysis}, explaining why cell performance is more susceptible to merging-induced interference. Nevertheless, ES-Merging alleviates this degradation more effectively than existing merging approaches.
 
\begin{table}[t!]
\centering

\begin{minipage}[t]{0.63\textwidth}
\captionof{table}{Performance comparison on single-modality tasks. We report the accuracy for each task with the average accuracy.}
\vspace{0.05in}
\centering
\setlength{\tabcolsep}{8pt}
\renewcommand{\arraystretch}{1.15}
\resizebox{\linewidth}{!}{%
\begin{tabular}{l c c c >{\columncolor{gray!15}}c}
\toprule
& \textbf{Molecule} & \textbf{Protein} & \textbf{Cell} & \textit{Avg.} \\
\midrule
Specialized MLLM & 75.68 & 85.25 & 77.84 & 79.59 \\
\midrule
Avg. Merging                       & 64.34 & 70.87 & 55.94 & 63.72 \\
TIES-Merging~\citep{tiesmerging}       & 45.67 & 62.30 & 46.90 & 51.62 \\
EMR-Merging~\citep{emrmerging}         & 61.22 & 61.03 & 50.13 & 57.46 \\
AdaMerging~\citep{adamerging}          & 61.33 & 73.52 & 53.06 & 62.64 \\
PCB-Merging~\citep{pcbmerging}         & 49.99 & 57.88 & 50.14 & 52.67 \\
Consensus Merging~\citep{consensusmerging} & 48.39 & 67.59 & 46.94 & 54.31 \\
UQ-Merge~\citep{uqmerge}               & 39.56 & 79.32 & 52.42 & 57.10 \\
RobustMerge~\citep{robustmerge} & 49.14 & 64.19 & 55.16 & 56.16 \\
LS-Merge~\citep{lsmerge}               & 48.31 & 59.14 & 49.13 & 52.19 \\
\cdashline{1-5}[0.5pt/1pt]\rule{-2pt}{2.5ex}
\textbf{ES-Merging (Ours)}             & \textbf{77.64} & \textbf{86.51} & \textbf{61.49} & \textbf{75.21} \\
\bottomrule
\end{tabular}
}
\label{tab:single_modality_binary_prediction}
\end{minipage}
\hfill
\begin{minipage}[t]{0.35\textwidth}
\vspace{0pt}
\centering

\captionof{table}{Total FLOPs until determining merged LoRA parameters.}
\vspace{0.05in}
\renewcommand{\arraystretch}{1.17}
\setlength{\tabcolsep}{5pt}
\resizebox{\linewidth}{!}{%
\begin{tabular}{lc}
\toprule
\textbf{Method} & \textbf{Total FLOPs} $\downarrow$ \\
\midrule
Avg. Merging + FT          & 891,117 \\
AdaMerging                 & 493,694 \\
\textbf{ES-Merging (Ours)} & \textbf{149,807} \\
\bottomrule
\end{tabular}
}
\label{tab:flops}
\vspace{0.05in}

\captionof{table}{Ablation studies on different merging coefficient types.}
\vspace{-0.05in}
\renewcommand{\arraystretch}{1.17}
\setlength{\tabcolsep}{8pt}
\resizebox{\linewidth}{!}{%
\begin{tabular}{lccc}
\toprule
\textbf{Task} & \textbf{L} & \textbf{E} & \textbf{L$\times$E} \\
\midrule
Mol-Prot    & 63.6 & 64.9 & \textbf{65.7} \\
Mol-Cell    & 85.2 & 86.7 & \textbf{87.4} \\
CYP Inhibit & 73.9 & 72.7 & \textbf{74.5} \\
CYP Subs.   & 57.1 & 60.5 & \textbf{61.9} \\
\bottomrule
\end{tabular}
}
\label{tab:ablation-coefficient-combined}

\end{minipage}

\end{table}

\subsection{Analysis}

\paragraph{ES-Merging as an Efficient and Explainable Alternative to Fine-Tuning}
We further compare ES-Merging with task-specific fine-tuning from the perspectives of qualitative response behavior and computational efficiency. As shown in Tables~\ref{tab:case_study_mpi} and~\ref{tab:case_study_cell} of Appendix, both ES-Merging and the task-specific fine-tuned model predict the correct labels in the analyzed cases. Interestingly, ES-Merging provides biological rationales that explain its predictions by connecting features and information across different modalities to the target task, whereas the task-specific fine-tuned model outputs only the final label, showing that ES-Merging can retain more explicit cross-modal reasoning behavior learned by each specialized MLLM, rather than producing task labels.

ES-Merging is also substantially more efficient than optimization-based alternatives. As shown in Table~\ref{tab:flops}, ES-Merging reduces the computational cost of determining the merged LoRA parameters by 5.95$\times$ compared to task-specific fine-tuning and by 3.30$\times$ compared to AdaMerging. This efficiency comes from the fact that ES-Merging estimates merging coefficients once from probe-induced embedding signals, whereas task-specific fine-tuning and AdaMerging require iterative gradient-based optimization. These observations suggest that ES-Merging provides a practical alternative to task-specific fine-tuning, offering explicit biological rationales and avoiding expensive training while achieving the best or comparable performance over diverse cross-modal reasoning tasks.

\paragraph{Ablation Study}
We further conduct an ablation study to see the effectiveness of each coefficient type. As shown in Table~\ref{tab:ablation-coefficient-combined}, using only one of the coefficient types already outperforms most merging baselines, indicating that embedding space signals provide salient and robust information for cross-modal merging compared to parameter-signal-based merging methods. Combining these two coefficients consistently yields the best performance across all task groups, suggesting that effective MLLM merging requires both coarse-grained layer selection and fine-grained parameter selection: the former prevents globally important layers from being diluted, while the latter avoids assigning uniform weights to all parameters within a layer.

\paragraph{Merging Coefficient Analysis}
We further visualize the layer-wise merging coefficients in Fig.~\ref{fig:layer_coeff} and the element-wise merging coefficients in Fig.~\ref{fig:coef_qkvo} of Appendix. The coefficients are distinctly distributed across different MLLMs, showing that modality-specific knowledge is not incorporated uniformly throughout the model. Additionally, we observe that even when the layer-wise coefficient is high (e.g., the third layer of the molecule LLM), the element-wise merging coefficients within that layer vary substantially, as shown in Fig.~\ref{fig:mol_element_coeff_layer3}. This within-layer heterogeneity indicates that a high layer-wise coefficient alone does not imply that all parameters in the layer should be merged with the same strength. Instead, ES-Merging assigns larger weights only to the parameter elements that are more sensitive to modality-specific representation shifts, preventing irrelevant or weakly specialized parameters from being over-emphasized. Therefore, modality specialization emerges at multiple levels of granularity: only a few parameter elements within important layers are primarily salient for modality specialization. Therefore, combining layer-wise global coefficients with element-wise local coefficients leads to the complementary merging of different granularities, achieving an accurate and robust merging of different modalities.

\begin{figure*}[!t]
  \centering
  \includegraphics[width=\textwidth]{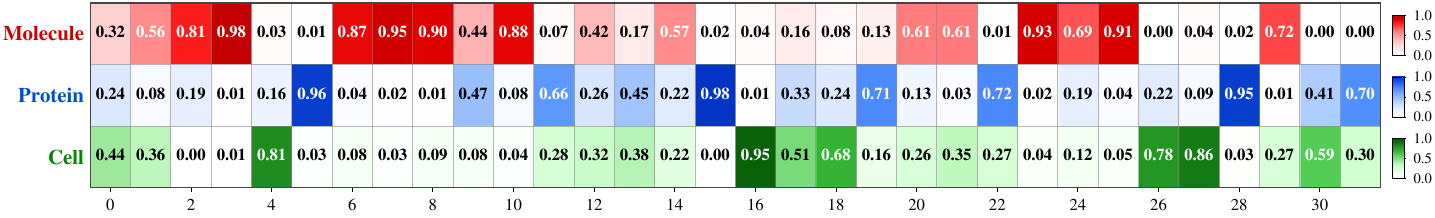}
  \vspace{-0.25in}
  \caption{Computed layer-wise merging coefficient visualization of each specialized MLLM. The $l$-th column corresponds to the merging coefficient of the $l$-th layer, $\alpha^l_{m_j}$.}
  \label{fig:layer_coeff}
\end{figure*}

\begin{table*}[t]
\centering

\begin{minipage}[t]{0.46\textwidth}
  \centering
  \vspace{0pt}
  \includegraphics[width=\linewidth]{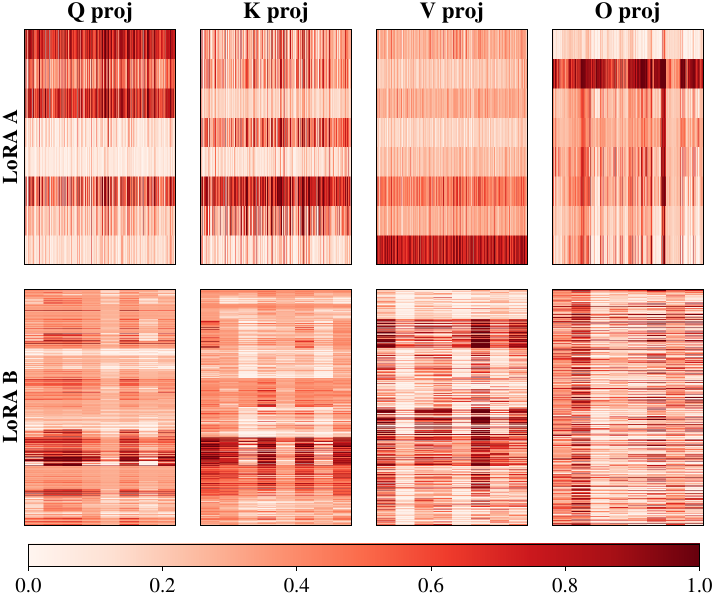}
  \vspace{-0.22in}
  \captionof{figure}{Element-wise merging coefficient visualization of the third layer in molecule LLM.}
  \label{fig:mol_element_coeff_layer3}
\end{minipage}
\hfill
\begin{minipage}[t]{0.52\textwidth}
  \centering
  \vspace{0pt}
  \captionof{table}{Performance comparison on the MUSIC-AVQA task across different modality combinations.}
  \vspace{-0.05in}
  \setlength{\tabcolsep}{3pt}
  \renewcommand{\arraystretch}{1.24}

  \resizebox{\linewidth}{!}{%
  \begin{tabular}{l c c c c c c c}
  \toprule
  \textbf{Method} & \textbf{V} & \textbf{I} & \textbf{A} & \textbf{V + I} & \textbf{V + A} & \textbf{I + A} & \textbf{V + I + A} \\
  \midrule
  \rowcolor{lightgreen}
  \multicolumn{8}{l}{\textit{Base LLM and Specialized MLLMs}} \\
  Vicuna-7B       & 25.74 & 30.52 & 18.89 & 22.07 & 26.96 & 24.65 & 30.98 \\
  VideoLLM        & 49.13 & 47.07 & 40.35 & 49.41 & 48.91 & 48.56 & 51.49 \\
  ImageLLM        & 49.67 & \underline{50.78} & 37.74 & 50.16 & 49.74 & 51.80 & 51.27 \\
  AudioLLM        & 30.84 & 43.07 & 28.28 & 29.55 & 29.61 & 37.69 & 30.75 \\

  \midrule

  \rowcolor{lightgreen}
  \multicolumn{8}{l}{\textit{Merging Methods}} \\
  Avg. Merging                       & 46.82 & 48.22 & 40.64 & 45.35 & 47.25 & 48.65 & 47.30 \\
  TIES-Merging~\citep{tiesmerging}   & 32.31 & 36.00 & 22.59 & 27.69 & 34.16 & 30.32 & 35.96 \\
  EMR-Merging~\citep{emrmerging}     & \underline{52.18} & 50.77 & 41.11 & \underline{52.28} & \underline{51.82} & 52.16 & \underline{53.58} \\
  AdaMerging~\citep{adamerging}      & 49.73 & 50.54 & \underline{43.18} & 49.63 & 50.21 & \underline{52.30} & 52.18 \\
  PCB-Merging~\citep{pcbmerging}     & 29.26 & 20.73 & 20.73 & 25.14 & 31.32 & 27.49 & 34.10 \\
  Consensus Merging~\citep{consensusmerging} & 32.41 & 21.96 & 21.96 & 27.91 & 33.73 & 29.18 & 35.77 \\
  UQ-Merge~\citep{uqmerge} & 50.77 & 48.68 & 37.74 & 50.41 & 50.65 & 49.52 & 51.62 \\
  RobustMerge~\citep{robustmerge}   & 51.64 & 50.05 & 42.09 & 51.80 & 51.61 & 51.03 & 52.93 \\ 
  \cdashline{1-8}[0.5pt/1pt]\rule{-2pt}{2.5ex}
  \textbf{ES-Merging (Ours)} & \textbf{53.69} & \textbf{53.11} & \textbf{44.10} & \textbf{53.76} & \textbf{53.06} & \textbf{53.89} & \textbf{53.59} \\
  \bottomrule
  \end{tabular}
  }
  \label{tab:music_avqa_merging_prediction}
\end{minipage}

\end{table*}

\paragraph{General Domain Application}

Although ES-Merging is primarily designed and evaluated for biological MLLM merging, its core principle is applicable to diverse domains: merging coefficients are estimated from embedding-space discrepancies rather than modality-specific assumptions. To examine whether this principle can extend beyond biological modalities, we additionally evaluate ES-Merging on the MUSIC-AVQA~\cite{musicavqa} benchmark which requires understanding visual and acoustic signals from video, image, and audio. As shown in Table~\ref{tab:music_avqa_merging_prediction}, ES-Merging achieves the best performance across all modality combinations, including single-, two-, and three-modality inputs. In particular, ES-Merging consistently outperforms parameter space merging baselines, suggesting that embedding space signals can also provide useful guidance for merging specialists in general multimodal domains. We also report additional experiments on AVQA in Appendix~\ref{app:sec:avqa}.

\begin{table}[t]
\centering
\caption{Accuracy (\%) on CYP2D6 inhibition under coefficient interventions.}
\label{tab:coefficient_contribution}
\small
\setlength{\tabcolsep}{5pt}
\renewcommand{\arraystretch}{1.1}
\begin{tabular}{lrrr}
\toprule
\textbf{Intervention} & \textbf{Original} & \textbf{Modified} & \textbf{Degradation (pp)} \\
\midrule
Remove high-coefficient layers & 79.2 & 69.7 & -9.5 \\
Remove high-coefficient elements & 77.6 & 62.3 & -15.3 \\
Reverse element-wise priority & 77.6 & 64.2 & -13.4 \\
\bottomrule
\end{tabular}
\end{table}

\paragraph{Merging Coefficient Contribution Validation}
To directly examine whether embedding-space discrepancy reflects actual task contribution, we selectively remove layers and elements assigned high molecule-specific coefficients. For the layer-wise-only variant, we remove molecule-expert parameters from the eight layers whose molecule coefficients exceed 0.8. For the element-wise-only variant, we remove the 6.255\% of all evaluated elements whose molecule coefficients exceed the same threshold. As shown in Table~\ref{tab:coefficient_contribution}, both interventions substantially degrade performance, supporting that high embedding-based coefficients identify task-relevant expert contributions. We further examine whether the relative ordering induced by the element-wise scores is meaningful. We take the reciprocal of the coefficients to downweight high-score elements and emphasize low-score elements, followed by normalization across experts. Reversing the coefficient priorities reduces accuracy by 13.4 percentage points. This provides additional evidence that high element-wise scores capture task-relevant and beneficial expert contributions rather than harmful perturbations.

\section{Conclusion}
We present ES-Merging, a novel MLLM merging framework that shifts model merging from relying on parameter space signals to leveraging embedding space signals. Our core insight is that input-aware representations encode rich modality-specific specialization, providing a more faithful basis for integrating modality-specialized MLLMs. Based on this observation, we introduce two complementary types of merging coefficients at different granularities: layer-wise global coefficients for capturing coarse-grained specialization and element-wise local coefficients for capturing fine-grained importance. Experiments on diverse biological tasks demonstrate that ES-Merging consistently improves cross-modal reasoning while better preserving single-modal expertise. Further analyses show that ES-Merging is an efficient and explainable alternative to task-specific fine-tuning and that embedding space signals can also generalize to non-biological multimodal merging scenarios. We believe that our work establishes embedding space signals as a principled and practical foundation for constructing unified MLLMs from heterogeneous modality-specialized experts.

\section*{Acknowledgments}
This work was supported by the Institute for Information \& communications Technology Planning \& Evaluation (IITP) grant funded by the Korea government (MSIT) (RS-2019-II190075, Artificial Intelligence Graduate School Program (KAIST)); the National Research Foundation of Korea (NRF) grant funded by the Korea government (MSIT) (No. RS-2023-00256259); the ``Advanced GPU Utilization Support Program'' funded by the Government of the Republic of Korea (Ministry of Science and ICT); the Korea Machine Learning Ledger Orchestration for Drug Discovery Project (K-MELLODDY), funded by the Ministry of Health \& Welfare and the Ministry of Science and ICT, Republic of Korea (grant number: RS-2024-00460870); the Institute of Information \& communications Technology Planning \& Evaluation (IITP) grant funded by the Korea government (MSIT) (No. RS-2022-II220713, Meta-learning Applicable to Real-world Problems); and the Institute of Information \& Communications Technology Planning \& Evaluation (IITP) grant funded by the Ministry of Science and ICT (MSIT) of the Republic of Korea in connection with the Global AI Frontier Lab International Collaborative Research (No. RS-2024-00469482 \& RS-2024-00509279).

\bibliography{reference}

\newpage
\appendix
\onecolumn
\begin{center}{\bf {\LARGE Appendix}}\end{center}
\paragraph{Organization} Appendix is organized as follows: In Section~\ref{sec:exp_details}, we provide detailed experimental settings, including datasets, baseline, implementation, and prompt setting details. In Section~\ref{sec:qual_results}, we qualitatively analyze generated responses. In Section~\ref{sec:detail_analysis}, we provide additional analysis of our merging method. In Section~\ref{app:sec:limitation_and_societal_impact}, we discuss the limitations and the societal impacts of our work.

\vspace{-11.5pt}
\section{Experiment Settings}
\label{sec:exp_details}
\vspace{-4.4pt}
\subsection{Dataset Details\label{sec:data_details}}
\vspace{-0.1in}
In this section, we provide detailed explanations of datasets used in our experiments. For all datasets, we leverage the splits that are officially provided in original repositories.
\vspace{-0.1in}
\begin{itemize}[itemsep=0.5mm, parsep=3pt, leftmargin=*]
    \item \textbf{BindingDB} is a drug-target interaction dataset derived from experimentally measured small molecule-protein binding data, typically filtered to human proteins, and used as a binary classification benchmark with 11,054 samples for predicting whether a drug-target pair interacts.
    \item \textbf{BioSNAP} is a drug-target interaction dataset derived from known associations between US-marketed drugs and their human protein targets, and is used as a binary classification benchmark with 6,058 samples for predicting whether a drug-target pair interacts.
    \item \textbf{Human} is a drug-target interaction dataset consisting of drug-human protein pairs with highly credible negative samples and used as a binary classification benchmark with 1,375 samples to predict whether a drug-target pair interacts.
    \item \textbf{DrugComb} is a drug combination-cell line interaction dataset derived from standardized and harmonized drug combination screening studies across various cancer cell lines, and is used as a binary classification benchmark with 3,631 samples for predicting whether a combination of two drugs produces a synergistic or antagonistic anticancer effect in a given cancer cell line.
    \item \textbf{GDSC2} is a drug-cell line interaction dataset derived from the Genomics of Drug Sensitivity in Cancer project, which screens over 1,000 genetically characterized human cancer cell lines with a wide range of anti-cancer therapeutics using a newer cell screening platform introduced in 2015, and is used as a binary classification benchmark with 843 samples for predicting whether a given cancer cell line is sensitive or resistant to a specific drug.
    
    \item \textbf{CYP1A2}, \textbf{CYP2C19}, \textbf{CYP2C9}, \textbf{CYP2D6}, and \textbf{CYP3A4 Inhibition} are binary classification benchmarks for predicting whether a given drug inhibits a specific cytochrome P450 enzyme isoform involved in drug metabolism, consisting of 2,516, 2,533, 2,418, 2,626, and 2,466 samples, respectively.
    \item \textbf{CYP2C9}, \textbf{CYP2D6}, and \textbf{CYP3A4 Substrate} are three binary classification benchmarks to predict whether a given drug is a substrate of a specific cytochrome P450 enzyme isoform involved in drug metabolism with 134, 133, and 134 samples, respectively.

    \item \textbf{PAMPA} is a molecule permeability dataset derived from parallel artificial membrane permeability assay measurements, and is used as a binary classification benchmark with 407 samples for predicting whether a given molecule exhibits high or low passive permeability across an artificial membrane.
    
    \item \textbf{Immuno-Virus} is a viral protein immunogenicity dataset compiled by ~\citet{venusvaccine}, derived from curated viral antigen records with experimentally annotated immunogenicity labels, and is used as a binary classification benchmark with 397 samples for predicting whether a given viral protein is a protective antigen or a non-protective antigen from its amino acid sequence.
    
    \item \textbf{PBMC} is a single-cell gene expression dataset derived from peripheral blood mononuclear cells profiled by single-cell RNA sequencing, and is used as a multi-class classification benchmark with 2,398 samples for predicting the cell type of a given cell from its ranked gene-expression profile.

    \item \textbf{MUSIC-AVQA}~\citep{musicavqa} is an audio-visual question answering benchmark over musical performance videos, requiring joint reasoning across visual, audio, and language modalities to answer questions about sounding objects, their attributes, and temporal-spatial associations, consisting of 9,124 samples. The questions encompass both binary and open-ended formats. 
    
\end{itemize}
\subsection{Baseline Details\label{sec:base_details}}

We evaluate our method on both the biological and general domains. For each domain, we adopt a distinct set of modality-specialized base models, while sharing six model merging baselines and three ablation variants of our proposed method.

\paragraph{Specialized Models for Biological Domain}
\vspace{-0.05in}
\begin{itemize}[itemsep=0.5mm, parsep=3pt, leftmargin=*]
    \item \textbf{LLaMA-3.1-8B-Instruct} \citep{llama3} is the shared base large language model for all three biological specialist models, providing the general-purpose instruction-following capabilities that are subsequently adapted via LoRA for each modality-specific task.
    \item \textbf{Mol-LLaMA} \citep{molllama} is a large molecular language model trained via multi-modal instruction tuning that integrates complementary 2D and 3D molecular encoders through a blending module, with a Q-Former projector and a LLaMA backbone fine-tuned with LoRA, providing general molecular understanding with explainability and reasoning capabilities across diverse molecular tasks.
    \item \textbf{Prot2Text-V2} \citep{prot2textv2} is a multi-modal sequence-to-text model that combines an ESM2-3B~\citep{esm2} protein sequence encoder with a LLaMA-3.1-8B-Instruct decoder via a nonlinear modality projector, using hybrid sequence-level contrastive alignment learning and instruction-based LoRA fine-tuning to generate rich functional descriptions of proteins directly from amino acid sequences.
    \item \textbf{Cell-o1} \citep{cello1} is a cell line-specialized model fine-tuned on transcriptomic omics data, capable of handling cell line-related drug response tasks but limited to single-modality inference without understanding molecule or protein.
    \item \textbf{Average Merging + Task-Specific Finetune} constructs a base model by averaging the LoRA weights of all three specialist models, and subsequently fine-tunes the LoRA parameters independently on each evaluation task by sampling 2,000 examples from the training dataset for 10 epochs with a batch size of 16.
\end{itemize}
    
\paragraph{Specialized Models for General Domain}
\vspace{-0.05in}
\begin{itemize}[itemsep=0.5mm, parsep=3pt, leftmargin=*]
    \item \textbf{Vicuna-7B-v1.5} \citep{vicuna} is the shared base large language model upon which all three general-domain specialist models are built, adapted via LoRA for each modality-specific task.
    \item \textbf{Audio LLM} is a multi-modal model that combines a BEATs~\citep{beats} audio encoder with a Vicuna-7B-v1.5 backbone via a Q-Former audio projector, and leverages LoRA-based fine-tuning that enables the model to capture acoustic information from speech, music, and environmental sounds.
    \item \textbf{Video LLM} is a multi-modal model that integrates a LanguageBind-Video~\citep{languagebind} video encoder with a Vicuna-7B-v1.5 backbone via a two-layer MLP projector with GELU~\citep{gleu} activation, and is fine-tuned with LoRA to perform temporal reasoning over short video clips.
    \item \textbf{Image LLM} is a multi-modal model that integrates a CLIP-ViT-L/14-336~\citep{clip} image encoder with a Vicuna-7B-v1.5 backbone via a two-layer MLP projector with GELU activation, enabling visual reasoning over single images through LoRA-based instruction tuning.
\end{itemize}

\paragraph{Common Model Merging Methods}
\vspace{-0.05in}
\begin{itemize}[itemsep=0.5mm, parsep=3pt, leftmargin=*]
    \item \textbf{Average Merging} directly averages the parameter elements of all specialist MLLMs parameters without any additional computation.
    \item \textbf{TIES-Merging} \citep{tiesmerging} resolves interference among task vectors through three steps, trim, elect sign, and disjoint merge, which prune low-magnitude parameters, resolve sign conflicts by selecting the dominant direction, and merge only the parameter-aligned subset.
    \item \textbf{AdaMerging} \citep{adamerging} learns merging coefficients for task vectors in a test-time adaptation manner, using the minimization of model output entropy on unlabeled test samples as a surrogate objective without relying on original training data. In our experiments, we adopt the layer-wise AdaMerging variant which independently learns a merging coefficient for each layer of every task vector, showing better performance compared to the task-wise AdaMerging.
    \item \textbf{EMR-Merging} \citep{emrmerging} is a tuning-free method that first elects a unified task vector by selecting the maximum absolute value of each parameter along the dominant sign direction, and then generates lightweight task-specific masks and rescalers to align the direction and magnitude of the unified model with each original specialist model at inference time.
    \item \textbf{PCB-Merging} \citep{pcbmerging} is a training-free method that constructs a parameter competition balancing matrix through intra-balancing, which measures parameter significance within individual tasks, and inter-balancing, which assesses parameter similarity across tasks, then drops low-scoring parameters and rescales the remaining ones to form the final merged model.
    \item \textbf{Consensus-Merging} \citep{consensusmerging}  identifies and removes two classes of detrimental parameters, selfish weights that are critical exclusively to a single task and catastrophic weights that are irrelevant to all tasks, and retains only the consensus parameters that contribute positively to multi-task fusion.
    \item \textbf{UQ-Merge} \citep{uqmerge} measures each model's prediction confidence via input perturbation-based uncertainty quantification, and incrementally merges models sorted in descending order of uncertainty, stopping when the merged model's uncertainty no longer decreases, thereby excluding over-confident models from the merging process.
    \item \textbf{RobustMerge} \citep{robustmerge} prunes low-magnitude parameters, adaptively amplifies small singular values through complementary scaling coefficients, and applies cross-task normalization, thus maintaining direction robustness in the low-rank space to mitigate parameter interference in a parameter-efficient manner.
    \item \textbf{LS-Merge} \citep{lsmerge} encodes model weights into a latent space via a transformer-based VAE, performs merging through linear interpolation in that space, and decodes back to parameters.
    \item \textbf{Layer-wise ES-Merging (Ours)} applies only the layer-wise global coefficient $\alpha^{l}_{m_j}$, derived from SWD-based embedding distribution shifts between the base and specialized models, as the merging coefficient.
    \item \textbf{Element-wise ES-Merging (Ours)} applies only the element-wise local coefficient $\beta^{l,n}_{m_j}$, derived from gradient-based parameter sensitivity scores with respect to fine-grained embedding distances, as the merging coefficient.
    \item \textbf{ES-Merging (Ours)} integrates both the global layer coefficient $\alpha^{l}_{m_i}$ and the local element coefficient $\beta^{l,n}_{m_i}$ into a unified merging coefficient $\lambda^{l,n}_{m_i}$ by multiplication and renormalization of elements in modalities.
\end{itemize}

\subsection{Implementation Details\label{sec:impl_details}}

\paragraph{LoRA Configuration} All specialized models adopt LoRA~\citep{hu2022lora} on the self-attention projection matrices ($\mathbf{W}_Q, \mathbf{W}_K, \mathbf{W}_V, \mathbf{W}_O$) and MLP projection layers (gate, up, and down projections) of every transformer block, with rank $r=8$ and scaling factor $\alpha=32$ for the biological specialized models, and rank $r=128$ and scaling factor $\alpha=256$ for the general-domain specialized models.

\paragraph{Details of ES-Merging} To construct the probe inputs, we randomly sample the 110 samples for each modality from a collection of train sets, constructing 330 samples in total. For the layer-wise merging coefficient computation, we leverage SWD with slice projection dim$=$1,024 and distance order $p{=}2.0$, then normalized via softmax with temperature $\tau{=}0.5$. For the element-wise merging coefficient computation, the temperature $\tau$ of the softmax is predefined as 0.5.

\paragraph{Existing Merging Methods} For all baseline methods, we follow the experimental settings reported in their respective original papers and official implementations. For LS-Merge \citep{lsmerge}, we train VAE with sequence length $16{,}384$, batch size $512$, learning rate $3\times10^{-4}$ with $500$ step warmup and cosine decay, and KL weight $10^{-4}$ with adaptive adjustment toward a target KL of $50$, for $1{,}000$ epochs. At merge time, model parameters are split into chunks of size $16{,}384$, encoded into the latent space via the VAE encoder, merged via uniform-weighted Euclidean mean of the posterior means $\mu$, and decoded back to parameter space.

\paragraph{LLM Inference}
To ensure reproducibility and reduce decoding variance, we use deterministic greedy decoding for all methods, without sampling. Specifically, at each decoding step, the model selects the token with the highest predicted probability.

\paragraph{Compute Resources} All experiments are conducted on NVIDIA B200 and H200. For the computation of merging coefficients, it takes one and a half hours, while the experiments on the instance-varying interaction prediction and the target-fixed functionality prediction take 25 and 7 hours.

\begin{figure*}[t]
    \centering
    \includegraphics[width=\textwidth]{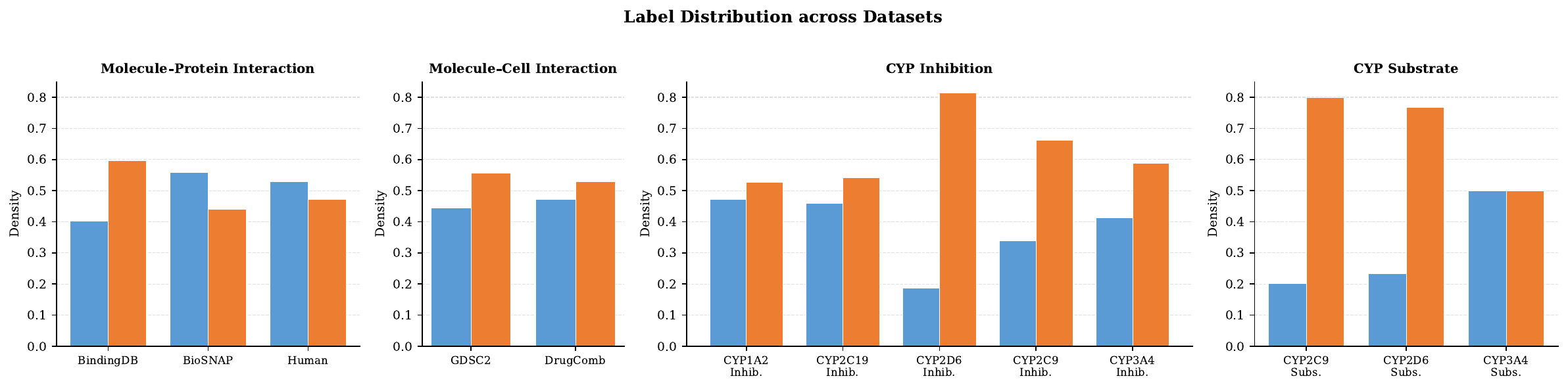}
    \caption{Label distribution across datasets. Each panel shows positive and negative sample counts for Molecule-Protein Interaction, CYP Inhibition, CYP Substrate, and Molecule-Cell Interaction tasks.}
    \label{fig:label_distribution}
\end{figure*}

\subsection{Evaluation Prompts\label{sec:prompt_setting}}
To impart the task-specific knowledge, we provide 5-shot examples from the training set with the prompt templates in 
Table~\ref{tab:prompt_DTI_CYP} and Table~\ref{tab:prompt-Cell}. 
The retrieval strategy is designed per task as follows:

\begin{itemize}[itemsep=0.5mm, parsep=3pt, leftmargin=*]
    \item \textbf{Molecule-Protein Interaction:} Training samples whose target protein sequence exactly matches the query are first collected. If more than five exact matches exist, the top-5 are selected by Tanimoto similarity~\citep{tanimoto} of Morgan fingerprints~\citep{fingerprint} between their molecules and the query molecule. If fewer than five are found, the remaining slots are filled with samples from proteins with the highest cosine similarity of protein embeddings from ESM2 to the query protein. Among candidates from the same similar protein, the one with the highest Tanimoto similarity to the query molecule is preferred.
    
    \item \textbf{Drug-Cell Interaction:} Training samples whose cell line shares an exact match on the top-50 expressed gene set are priorly collected. If fewer than five are found, additional samples are drawn from cell lines with the highest Jaccard similarity on gene sets. When multiple candidates exist from the same cell line, the one with the highest Tanimoto similarity to the query molecule is selected. For DrugComb, molecule similarity is computed as the average Tanimoto similarity across both drugs.
    
    \item \textbf{CYP Inhibition/Substrate:} Since all samples share the same CYP enzyme target, protein-level filtering is unnecessary. The top-5 examples are selected solely by Tanimoto similarity between the training molecules and the query molecule.

    \item \textbf{PAMPA/Immuno-Virus/PBMC:} As these single-modality tasks correspond to out-of-distribution datasets that the specialist models have not been trained on, we employ a task-grounded zero-shot prompt that consists of (i) a concise information of the task, and (ii) explanations of the analytical factors that the model should consider when reasoning toward the answer. The model is then instructed to reason step-by-step over these factors before producing the final answer.

    \item \textbf{MUSIC-AVQA:} Following~\citet{chen2024model}, we issue the same questions across multiple input-modality configurations of $V$, $I$, $A$, $V{+}I$, $V{+}A$, $I{+}A$, and $V{+}I{+}A$, where $V$, $I$, and $A$ denote video, image, and audio inputs, respectively. For each configuration, only the corresponding modalities are inserted into the prompt while the question text and answer format remain identical.

    \item \textbf{AVQA:} We evaluate four-option multiple-choice questions about real-world scenes, objects, actions, and sound-producing events~\citep{yang2022avqa} using the same seven input-modality configurations as MUSIC-AVQA. Each question has one correct answer and three distractors. For each configuration, the merged model receives the question together with only the selected video, image, and/or audio inputs.

\end{itemize}
\subsection{Evaluation Metric}
As shown in Figure~\ref{fig:label_distribution}, while some datasets exhibit a relatively balanced distribution between positive and negative samples, others such as BindingDB and CYP2D6 Inhibition display a pronounced class imbalance. In particular, the CYP Substrate task contains only approximately 130 samples per dataset, posing a critical challenge of absolute data scarcity. In such imbalanced and low resource settings, relying solely on accuracy as an evaluation metric can be misleading, as a model that predominantly predicts the majority class may still achieve inflated scores without genuinely capturing minority-class patterns. Therefore, we leverage the macro-F1 that addresses this limitation by computing the F1 score for each class independently and averaging them with equal weight regardless of class frequency, thereby providing a fair assessment of predictive performance across all classes.

\section{Qualitative Analysis}
\label{sec:qual_results}
In this section, we qualitatively compare the generated responses of each task-specific fine-tuned model with those of ES-Merging, as shown in Table~\ref{tab:case_study_mpi} and Table~\ref{tab:case_study_cell}.
\begin{table*}[!ht]
\centering
\small
\renewcommand{\arraystretch}{1.13}
\setlength{\tabcolsep}{4pt}
\begin{tabular}{p{0.95\textwidth}}
\toprule
\begin{center}
    \includegraphics[width=\linewidth]{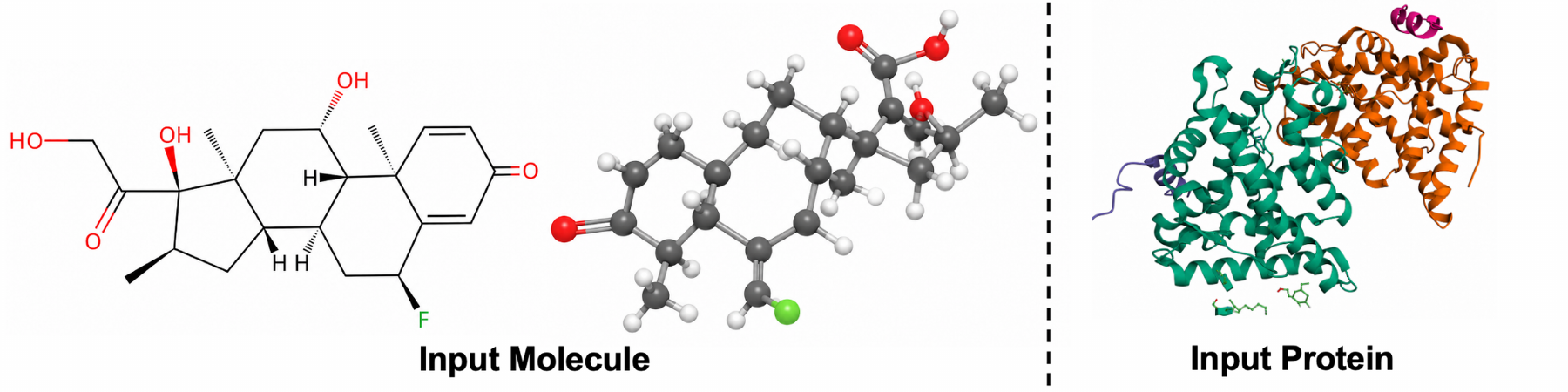}
\end{center} \\
\midrule
\textbf{Molecule Name:} Paramethasone \\
\midrule
\textbf{SMILES:} \seqsplit{C[C@@H]1C[C@H]2[C@@H]3C[C@@H](C4=CC(=O)C=C[C@@]4([C@H]3[C@H](C[C@@]2([C@]1(C(=O)CO)O)C)O)C)F} \\
\midrule
\textbf{Protein:} Glucocorticoid receptor \\
\midrule
\textbf{Protein Sequence:} \seqsplit{MDSKESLTPGREENPSSVLAQERGDVMDFYKTLRGGATVKVSASSPSLAVASQSDSKQRRLLVDFPKGSVSNAQQPDLSKAVSLSMGLYMGETETKVMGNDLGFPQQGQISLSSGETDLKLLEESIANLNRSTSVPENPKSSASTAVSAAPTEKEFPKTHSDVSSEQQHLKGQTGTNGGNVKLYTTDQSTFDILQDLEFSSGSPGKETNESPWRSDLLIDENCLLSPLAGEDDSFLLEGNSNEDCKPLILPDTKPKIKDNGDLVLSSPSNVTLPQVKTEKEDFIELCTPGVIKQEKLGTVYCQASFPGANIIGNKMSAISVHGVSTSGGQMYHYDMNTASLSQQQDQKPIFNVIPPIPVGSENWNRCQGSGDDNLTSLGTLNFPGRTVFSNGYSSPSMRPDVSSPPSSSSTATTGPPPKLCLVCSDEASGCHYGVLTCGSCKVFFKRAVEGQHNYLCAGRNDCIIDKIRRKNCPACRYRKCLQAGMNLEARKTKKKIKGIQQATTGVSQETSENPGNKTIVPATLPQLTPTLVSLLEVIEPEVLYAGYDSSVPDSTWRIMTTLNMLGGRQVIAAVKWAKAIPGFRNLHLDDQMTLLQYSWMFLMAFALGWRSYRQSSANLLCFAPDLIINEQRMTLPCMYDQCKHMLYVSSELHRLQVSYEEYLCMKTLLLLSSVPKDGLKSQELFDEIRMTYIKELGKAIVKREGNSSQNWQRFYQLTKLLDSMHEVVENLLNYCFQTFLDKTMSIEFPEMLAEIITNQIPKYSNGNIKKLLFHQK} \\
\midrule
\textbf{User:} Determine whether the given molecule interacts with the protein by following the example. \\
\midrule
\textbf{Ground Truth:} Interacts \\
\midrule
\textbf{ES-Merging (Ours)} \\ 
To determine whether the given molecule interacts with the protein, we need to analyze the molecular structure and properties of both the protein and the molecule.\\ 
\textcolor{blue}{The protein is a glucocorticoid receptor, which is a type of nuclear receptor} that plays a crucial role in regulating gene expression in response to glucocorticoids. 
\textcolor{blue}{The protein has a specific binding site for glucocorticoids, and it is activated by the binding of these hormones.}\\ 
\textcolor{blue}{The molecule is a synthetic glucocorticoid}, which is a type of steroid hormone that is designed to mimic the effects of natural glucocorticoids. It has a similar molecular structure to natural glucocorticoids and is known to bind to the glucocorticoid receptor.\\
Based on the molecular structure and properties of both the protein and the molecule, \textcolor{blue}{it is likely that the molecule interacts with the protein.} The molecule is designed to bind to the glucocorticoid receptor, and it has a similar structure to natural glucocorticoids, which are known to interact with this protein.\\
Therefore, the final answer is:\\
\textcolor{blue}{Final answer: Interacts} \\
\midrule
\textbf{Models with Task-Specific Finetuning} \\ \textcolor{blue}{Final answer: Interacts} \\
\bottomrule
\end{tabular}
\caption{Case study on Molecule-Protein Interaction prediction.}
\vspace{-0.05in}
\label{tab:case_study_mpi}
\end{table*}
\subsection{Molecule-Protein Interaction Prediction}
Table~\ref{tab:case_study_mpi} presents a generated response on the molecule-protein interaction prediction task from the BioSNAP dataset, where the molecule is paramethasone and the target protein is the glucocorticoid receptor. Although both ES-Merging and the task-specific fine-tuning model correctly predict the label, the two responses differ substantially in quality. ES-Merging correctly identifies the target as the glucocorticoid receptor, a nuclear receptor that regulates gene expression in response to glucocorticoids, and recognizes the molecule as a fluorinated steroid with glucocorticoid properties. Based on these complementary observations, it reasons that the molecule is likely to interact with the glucocorticoid receptor, which is consistent with paramethasone being annotated as a glucocorticoid receptor agonist targeting NR3C1. This reasoning process suggests that ES-Merging successfully integrates chemical structure-level knowledge acquired from the molecule expert model with biological function-level knowledge acquired from the protein expert model through embedding-space-signal-based merging. In contrast, the task-specific fine-tuned model outputs only the label ``Interacts'' without providing any biological or chemical rationale. Because it was trained on instruction data containing labels only, its predictions are less interpretable, whereas ES-Merging integrates modality-expert knowledge to produce interpretable reasoning even for unseen cross-modal tasks.

\begin{table*}[!ht]
\centering
\small
\renewcommand{\arraystretch}{1.3}
\setlength{\tabcolsep}{6pt}
\begin{tabular}{p{0.95\textwidth}}
\toprule
\begin{center}
    \includegraphics[width=\linewidth]{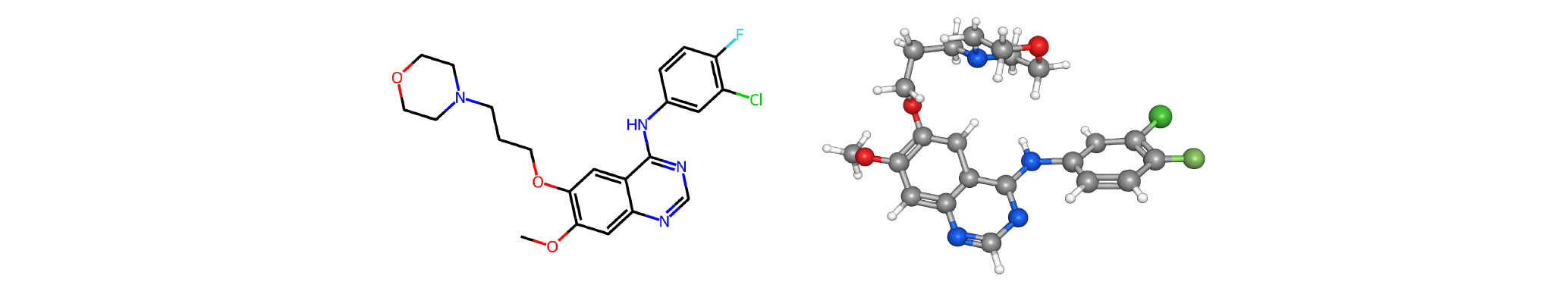}
\end{center} \\
\midrule
\textbf{Drug Name:} Gefitinib \\
\midrule

\textbf{SMILES:} COC1=C(C=C2C(=C1)N=CN=C2NC3=CC(=C(C=C3)F)Cl)OCCCN4CCOCC4 \\
\midrule

\textbf{Cell Line:} OVCA420 \\
\midrule

\textbf{Top 50 Genes:} RPL35, FTL, MT-ND4, GAPDH, AKR1B10, UBC, RPS6, RPL24, EIF1, GNB2L1, SPP1, RPS11, AKR1C1, RPL19, GNG5, AKR1B10P1, ACTG1, MT-CO3, RPL5, FAU, RPS15A, TXN, RPL8, MIF, LDHB, KRT19, EIF4H, RPL11, ALDH1A1, MGST1, RPS12, CLIC1, SNRPD2, RPL30, H2AFX, NQO1, PRDX1, RPS5, RPLP1, H2AFZ, TUBB4B, FAUP1, NACA, RPL14, HIST1H2BK, MYL12B, SHFM1P1, HINT1P1, RPL27, HIST1H4C \\
\midrule
\textbf{User:} Determine whether the given cell is sensitive or resistant to the given molecule based on the ranked gene-expression list. \\
\midrule
\textbf{Ground Truth:} Sensitive \\
\midrule

\textbf{ES-Merging (Ours)} \\ The given molecule is 4-(3-fluoro-4-methoxyphenyl)-7-methyl-6-oxo-1,4-dihydroquinazoline-3-carboximidamide, which is a \textcolor{blue}{quinazoline derivative}. \textcolor{blue}{This type of molecule is known to be a potent inhibitor of the epidermal growth factor receptor (EGFR) tyrosine kinase}. The cell line provided is a \textcolor{blue}{cancer cell}, and the presence of genes such as \textcolor{blue}{RPS6, EIF1, and GNB2L1} in the list suggests that the cell is actively proliferating and may be dependent on \textcolor{blue}{EGFR signaling} for growth and survival. Given the presence of these genes and the fact that \textcolor{blue}{the molecule is an EGFR inhibitor}, it is likely that the cell is sensitive to the molecule. \\ \textcolor{blue}{Final answer: Sensitive} \\
\midrule
\textbf{Models with Task-Specific Finetuning} \\ \textcolor{blue}{Final answer: Sensitive} \\

\bottomrule
\end{tabular}
\caption{Case study on Molecule-Cell Interaction prediction.}
\label{tab:case_study_cell}
\end{table*}
\subsection{Molecule-Cell Interaction Prediction}
Table~\ref{tab:case_study_cell} presents a qualitative analysis of the Drug-Cell Interaction prediction task from the GDSC2 dataset, where the drug is Gefitinib, and the cell line is OVCA420. ES-Merging first identifies the given molecule as a quinazoline derivative and correctly recognizes it as a potent inhibitor of the epidermal growth factor receptor (EGFR) tyrosine kinase. It also accurately identifies the given cell line as a cancer cell, then focuses on genes such as RPS6, EIF1, and GNB2L1 in the ranked gene-expression list to interpret the cell as actively proliferating and potentially dependent on EGFR signaling. On the basis of this, ES-Merging correctly recognizes that the drug's mechanism of action is effective against the given cell. This reasoning chain spanning from molecular substructure to drug class, target pathway, cell type identification, and cell-level interpretation demonstrates that the structural knowledge from the molecule expert model and the transcriptomic knowledge from the cell expert model are functionally integrated through ES-Merging, while the model with task-specific finetuning again outputs only the label ``Sensitive'' without any biological rationale.

\section{Further Analyses}
\label{sec:detail_analysis}
We provide further analyses regarding the merging coefficients, single-modal tasks, ablation study, temperature parameter $\tau$, probe prompt type, probe input sensitivity, embedding visualizations, and additional AVQA evaluation.
\subsection{Merging Coefficients}
\subsubsection{Layer-wise Merging Coefficients}
\begin{table}[!ht]
\centering
\small
\setlength{\tabcolsep}{8pt}
\renewcommand{\arraystretch}{1.2}
\begin{tabular}{lcccc} 
\toprule
Size & Human & GDSC2 & CYP2C9 Inh. & CYP2C9 Sub.  \\
\midrule
32    & 60.7 & 90.6  & 68.2 & 53.7 \\
256   & 62.0 & 93.1 & 68.9 & 48.5 \\
1024  & \textbf{62.0} & \textbf{94.1} & \textbf{72.5} & \textbf{64.2} \\
\bottomrule
\end{tabular}
\vspace{0.05in}
\caption{Additional studies on projection size of SWD for ES-Merging across Molecule-Protein and Molecule-Cell Interaction tasks.}
\label{tab:ablation-scale}
\end{table}

\paragraph{Projection Dimension Size of SWD} When computing SWD for the layer-wise merging coefficients, we vary projection dimension sizes on representative datasets for each task, and derived merging coefficients accordingly. As shown in Table~\ref{tab:ablation-scale}, performance consistently improves as the projection size increases, with 1024 dimensions achieving the best results. This is because a larger number of projections enables a more precise approximation of the distributional differences in high-dimensional space, leading to more accurate computation of the layer-wise merging coefficients. Based on this finding, we set the SWD projection size to 1024 in our main experiments.

\subsubsection{Element-wise Merging Coefficients}
\begin{figure*}[t!]
    \centering
    \includegraphics[width=\textwidth]{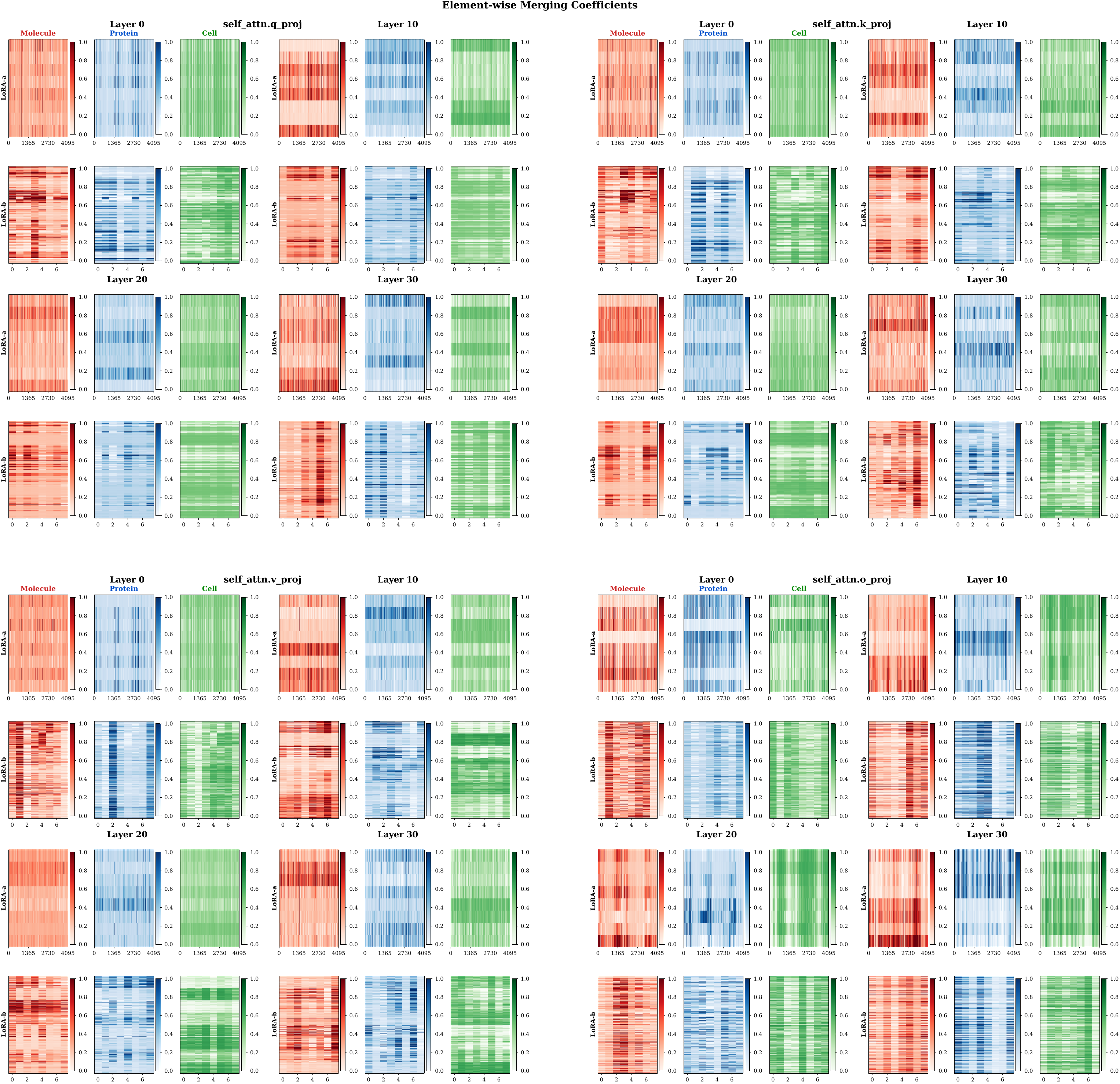}
    \caption{Element-wise merging coefficients of the LoRA parameters for \texttt{self\_attn.q/k/v/o\_proj} modules at layers 0, 10, 20, and 30. Rows correspond to LoRA-A and LoRA-B, and columns correspond to Molecule, Protein, and Cell modalities.}
    \label{fig:coef_qkvo}
\end{figure*}

The coefficient patterns differ across q/k/v/o\_proj modules even within the same layer. As shown in Figure~\ref{fig:coef_qkvo}, the q/k/v projection modules and the o projection module emphasize different elements even at Layer 0. Within the same module, the coefficient distributions between LoRA A and LoRA B exhibit distinct patterns. In particular, at Layer 0 of the q projection module, LoRA A shows relatively balanced coefficients across all modalities, whereas LoRA B contains regions where molecule and protein are more prominent. On the other hand, coefficient distributions are far from uniform, indicating that layer-wise global coefficients alone cannot capture the fine-grained specialization differences and supporting the necessity of element-wise local coefficients.

\subsubsection{Final Coefficients of ES-Merging}
\begin{figure*}[t!]
    \centering
    \captionsetup{justification=centering, singlelinecheck=true, width=\textwidth}
    \begin{minipage}[t]{0.48\linewidth}
        \centering
        \includegraphics[width=\textwidth]{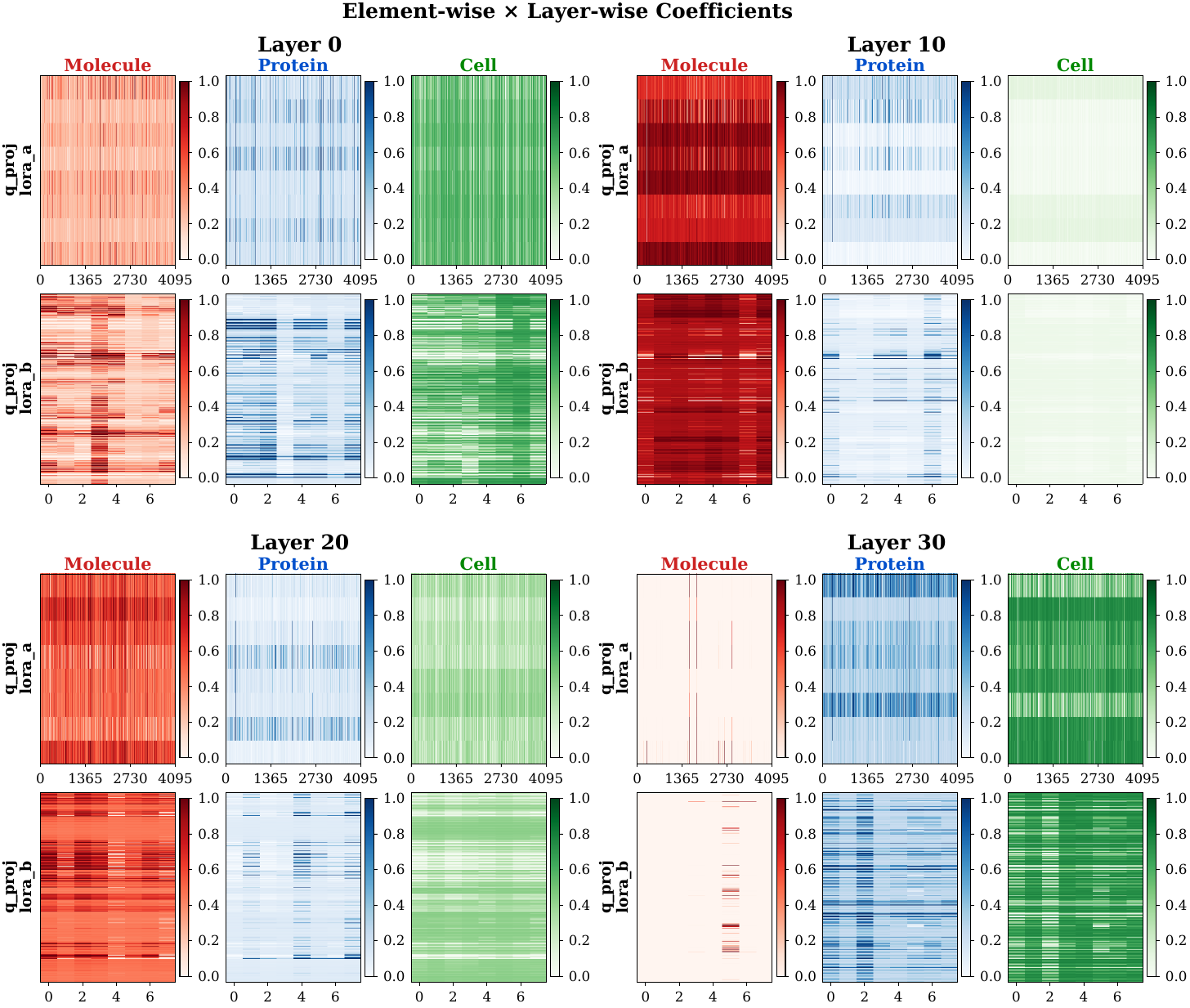}
        \vspace{-0.1in}
        \subcaption{q projection}
        \label{fig:mixed_coeff_q}
    \end{minipage}
    \hfill
    \begin{minipage}[t]{0.48\linewidth}
        \centering
        \includegraphics[width=\textwidth]{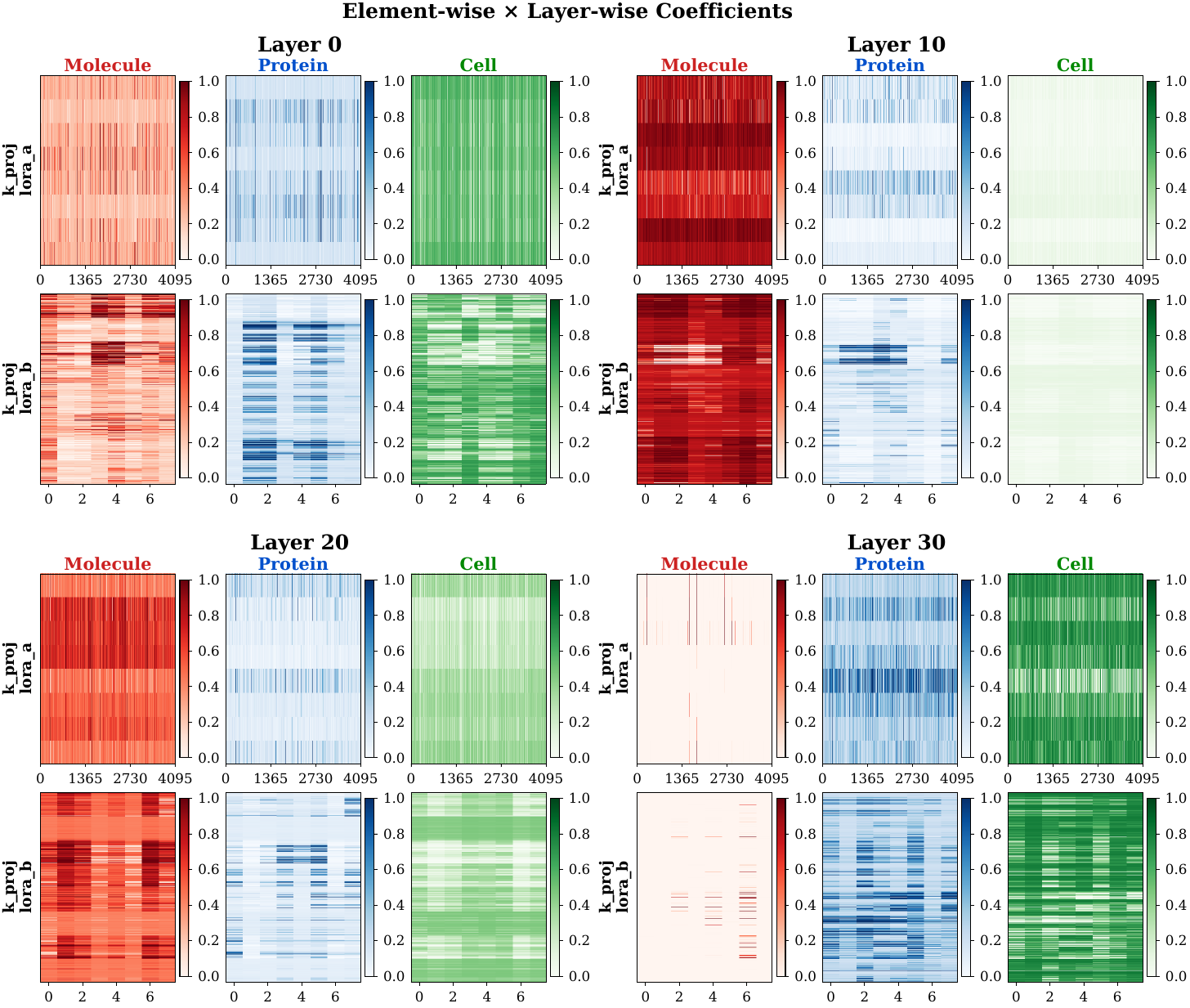}
        \vspace{-0.1in}
        \subcaption{k projection}
        \label{fig:mixed_coeff_k}
    \end{minipage}

    \vspace{6pt}

    \begin{minipage}[t]{0.48\linewidth}
        \centering
        \includegraphics[width=\textwidth]{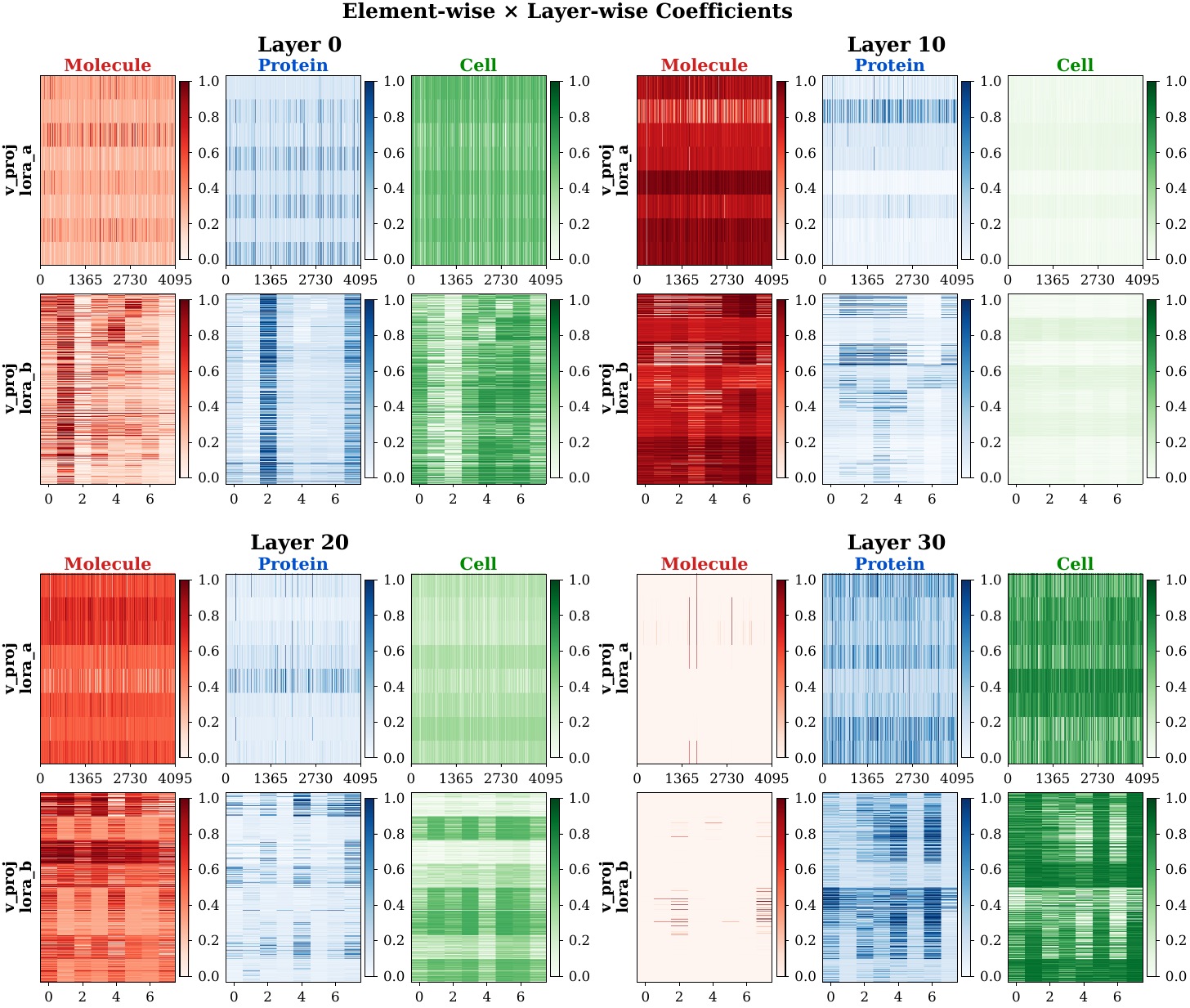}
        \vspace{-0.1in}
        \subcaption{v projection}
        \label{fig:mixed_coeff_v}
    \end{minipage}
    \hfill
    \begin{minipage}[t]{0.48\linewidth}
        \centering
        \includegraphics[width=\textwidth]{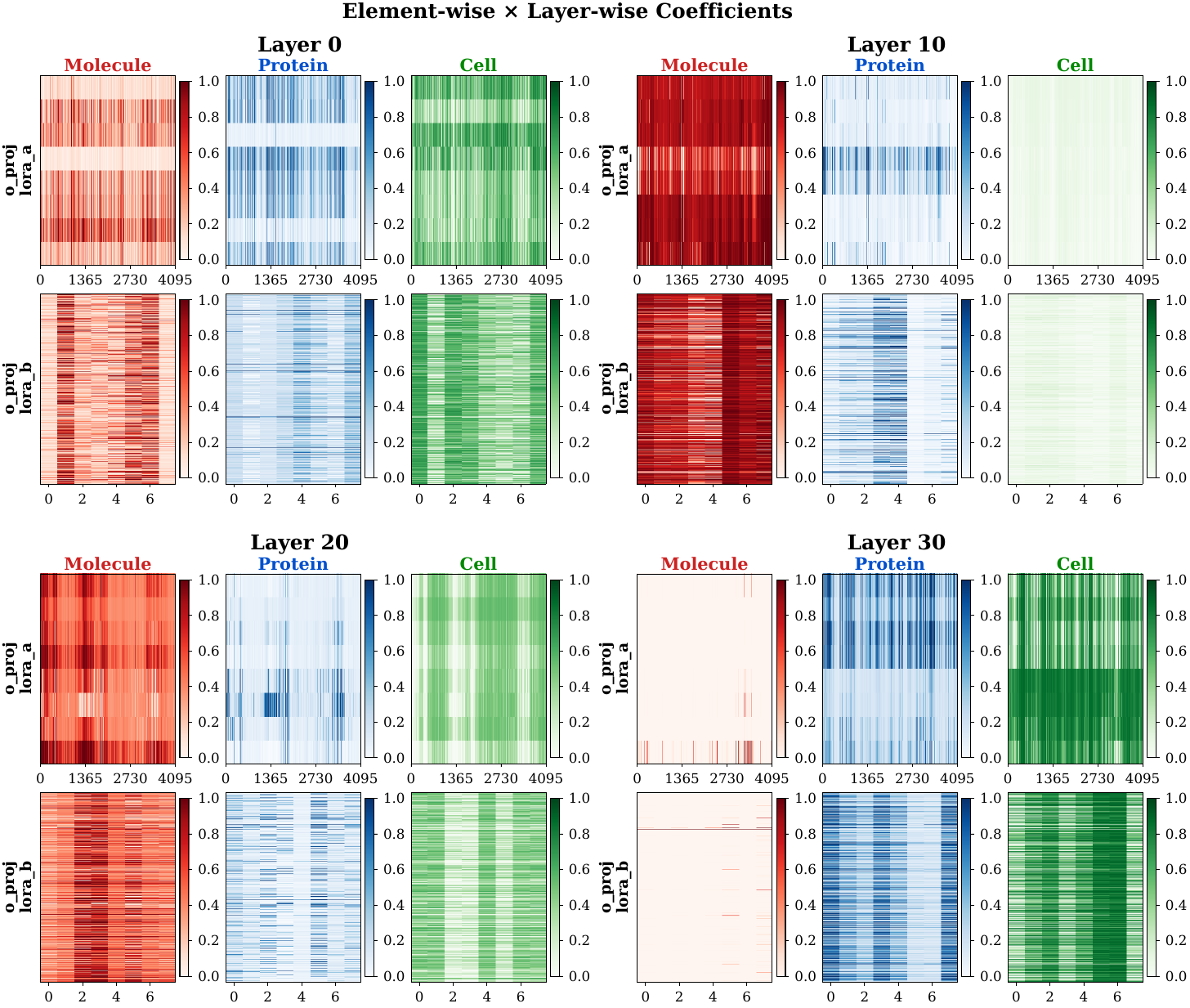}
        \vspace{-0.1in}
        \subcaption{o projection}
        \label{fig:mixed_coeff_o}
    \end{minipage}

    \vspace{-6pt}
    \caption{Element $\times$ Layer merging coefficients of the LoRA parameters for \texttt{self\_attn.q/k/v/o\_proj} modules at Layers 0, 10, 20, and 30, respectively.}
    \label{fig:mixed_qkvo}
\end{figure*}

The layer-wise global coefficient $\alpha^{l}_{m_i}$ and the element-wise local coefficient $\beta^{l,n}_{m_i}$ focus on different regions. The layer-wise coefficient captures coarse-grained distributional shifts in the embedding space, while the element-wise coefficient reflects fine-grained importance at the individual parameter level. As discussed in Section~\ref{sec:mixing-coefficients}, by multiplying these two coefficients and normalizing, regions where both coefficients assign high importance are amplified, whereas regions where only one side is high and the other is low are suppressed. This enables merging that simultaneously incorporates global layer-level specialization signals and local element-level specialization signals. As shown in Figure~\ref{fig:mixed_qkvo}, compared to the element-wise-only coefficient distributions, the combined results demonstrate that the overall scale of coefficients is adjusted according to the global importance of each layer, while the fine-grained element-level patterns are preserved.

\par\medskip
\noindent\begin{minipage}{\linewidth}
\centering
\captionof{table}{Accuracy (\%) on CYP2D6 inhibition with different modality aggregation strategies for coefficient estimation. Each column uses only the indicated coefficient type.}
\label{tab:coefficient_aggregation}
\normalsize
\setlength{\tabcolsep}{6pt}
\renewcommand{\arraystretch}{1.15}
\begin{tabular}{lcc}
\toprule
\textbf{Scoring strategy} & \shortstack{\textbf{Layer-wise}\\\textbf{only}} & \shortstack{\textbf{Element-wise}\\\textbf{only}} \\
\midrule
All modalities & 79.2 & 77.6 \\
Matched modality only & 78.3 & 71.2 \\
\bottomrule
\end{tabular}
\end{minipage}\par

\subsubsection{Modality Aggregation for Coefficient Estimation}
\label{app:sec:coefficient_aggregation}

We compare two strategies for aggregating the modality-conditioned signals used to estimate merging coefficients. Our default strategy equally aggregates signals from molecule, protein, and cell tokens for each expert. The matched-modality strategy uses only signals from the modality corresponding to each expert, e.g., molecule-token signals for the molecule expert. We evaluate both strategies separately in the layer-wise-only and element-wise-only variants on CYP2D6 inhibition. As shown in Table~\ref{tab:coefficient_aggregation}, relying only on the matched-modality signal underperforms all-modality aggregation, especially in the element-wise setting. This suggests that a single modality signal is insufficient for stable fine-grained attribution and supports jointly considering responses across all modalities to capture complementary contribution signals beyond modality-specific scoring.

\par\medskip
\noindent\begin{minipage}{\linewidth}
\centering
\small
\setlength{\tabcolsep}{8pt}
\renewcommand{\arraystretch}{1.2}
\begin{tabular}{l c c c}
\toprule
& \textbf{Molecule} & \textbf{Protein} & \textbf{Cell} \\
\midrule
Mol-LLaMA~\citep{molllama}         & 75.68 & 88.02 & 0.00 \\
Prot2Text-V2~\citep{prot2textv2}   & 39.80 & 85.25 & 0.00 \\
Cell-o1~\citep{cello1}             & 60.89 & 77.81 & 77.84 \\
\bottomrule
\end{tabular}
\vspace{-0.04in}
\captionof{table}{Performance comparison of different modality-specialized LLMs, evaluating their knowledge and capabilities not only on tasks within their target modality but also across other modalities. We report accuracy across each subset.}
\vspace{0.1in}
\label{tab:single_modality_specialized}
\end{minipage}\par

\subsection{Detailed Analysis of Single-modal Knowledge Preservation~\label{app:sec:single_modal_detailed_analysis}}

Table~\ref{tab:single_modality_specialized} reports whether each modality-specialized LLM retains a certain level of knowledge not only for its target modality but also for tasks in other modalities. We observe that Mol-LLaMA and Prot2Text-V2 attain zero accuracy on Cell tasks, whereas all specialized models possess a certain degree of cross-modal knowledge on the Molecule and Protein tasks. This distribution directly explains the results in Table~\ref{tab:single_modality_binary_prediction}. The drop in Cell performance compared to Cell-o1 alone arises because merging three models inevitably incurs a certain amount of information loss when the target model is combined with others that hold little to no knowledge of the corresponding task. Despite this constraint, ES-Merging preserves Cell knowledge most effectively among all merging methods and outperforms every baseline by a clear margin. In contrast, for the Molecule and Protein tasks, the non-target specialized models also encode a certain level of knowledge about each other's modalities, allowing ES-Merging to effectively integrate these complementary signals and even surpass the original modality-specialized MLLMs. These results confirm that ES-Merging selectively preserves and amplifies modality-relevant knowledge when it is present across the source models, while still retaining as much of that knowledge as possible even when it resides in only a subset of them.

\subsection{Details of Ablation Study}
\begin{table*}[!ht]
\centering
\small
\setlength{\tabcolsep}{2pt}
\renewcommand{\arraystretch}{1.2}
\resizebox{\textwidth}{!}{%
\footnotesize
\begin{tabular}{l ccc a cc a ccccc a ccc a}
\toprule
& \multicolumn{4}{c}{\textbf{Molecule-Protein Interaction}} & \multicolumn{3}{c}{\textbf{Molecule-Cell Interaction}} & \multicolumn{6}{c}{\textbf{CYP Inhibition}} & \multicolumn{4}{c}{\textbf{CYP Substrate}} \\
\cmidrule(l{2pt}r{2pt}){2-5} \cmidrule(l{2pt}r{2pt}){6-8} \cmidrule(l{2pt}r{2pt}){9-14} \cmidrule(l{2pt}r{2pt}){15-17}
Coefficient Type & BindingDB & BioSNAP & Human & \textit{Avg.} & DrugComb & GDSC2 & \textit{Avg.} & CYP1A2 & CYP2C19 & CYP2C9 & CYP2D6 & CYP3A4 & \textit{Avg.} & CYP2C9 & CYP2D6 & CYP3A4 & \textit{Avg.} \\
\midrule
Layer-wise              & 65.5 & \underline{68.2} & 57.0 & 63.6 & \underline{80.1} & \underline{90.2} & \underline{85.2} & \underline{76.3} & \textbf{71.4} & \underline{72.3} & \underline{79.2} & \underline{70.2} & \underline{73.9} & 61.2 & 54.9 & 55.2 & 57.1 \\
Element-wise   & \textbf{66.5} & 66.6 & \underline{61.4} & \underline{64.9} & 79.3 & \textbf{94.1} & 86.7 & 76.0 & 69.1 & 70.7 & 77.6 & 69.8 & 72.7 & \textbf{65.7} & \underline{57.6} & \underline{58.2} & \underline{60.5} \\
\textbf{Layer $\times$ Element Mixed} & \underline{66.0} & \textbf{69.1} & \textbf{62.0} &  \textbf{65.7} & \textbf{80.7} & \textbf{94.1} & \textbf{87.4} & \textbf{77.4}& \underline{70.6}& \textbf{72.5}& \textbf{80.7}& \textbf{71.3}& \textbf{74.5}& \underline{64.2}& \textbf{60.9}& \textbf{60.5}& \textbf{61.9}\\
\bottomrule
\end{tabular}%
}

\vspace{-0.05in}
\caption{Full results of the ablation studies on merging coefficient. We report accuracy across all datasets and their averages. \textbf{Bold} indicates the best and \underline{underline} indicates the second best.}
\label{tab:ablation-coefficient}
\end{table*}
Table~\ref{tab:ablation-coefficient} reports the detailed per-dataset results corresponding to the averages of each task group in Table~\ref{tab:ablation-coefficient-combined}. The layer-wise ES-Merging and element-wise ES-Merging each dominate on different datasets. 
Despite this dataset-level variations, Layer$\times$Element ES-Merging consistently achieves the best or comparable performance across all individual datasets. These results confirm that the two coefficients capture specialization at different granularities depending on task and dataset characteristics, and their combination compensates for the weaknesses of either side, yielding consistent and robust merging performance at the individual dataset level as well.

\begin{figure*}[!t]
    \centering
    \includegraphics[width=0.5\linewidth]{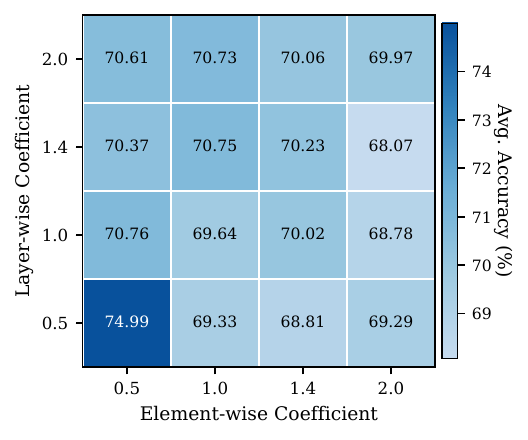}
    \caption{Performance with diverse temperatures $\tau$ for the layer-wise and element-wise coefficients in ES-Merging. Each cell reports the average accuracy (\%) across CYP2D6 Inhibition, CYP2D6 Substrate, Human, and GDSC2.}
    \label{fig:search_heatmap}
\end{figure*}

\subsection{Effect of Temperature parameter $\tau$ for Controlling Softmax}

The temperature parameter $\tau$ controls the sharpness of the softmax normalization used to convert embedding-space importance scores into merging coefficients. 
To see the effect of $\tau$, we compare the performance over $\tau_{\text{layer}}, \tau_{\text{elem}} \in \{0.5, 1.0, 1.4, 2.0\}$ by the average accuracy across four representative datasets: CYP2D6 Inhibition, CYP2D6 Substrate, Human, and GDSC2. 
As shown in Figure~\ref{fig:search_heatmap}, the best performance is achieved when both temperatures are set to 0.5. Increasing either temperature generally degrades performance, indicating that overly smooth coefficient distributions weaken the modality-specific signals captured from the embedding space.

\subsection{Probe Prompt Type}

\paragraph{Probe Prompt Settings} To compute the embedding-signal based merging coefficients, we feed a probe prompt that contains placeholder tokens for each modality into each expert model and extract their hidden representations. We consider two variants of the probe prompt. In the \textit{concatenated prompt} setting, the placeholders for all three modalities are packed into a single input sequence, so that the three modality signals are exposed to every expert model within the same context (Figure~\ref{fig:prompt-template}). In contrast, in the \textit{separated prompt} setting, we construct three modality-specific prompts so that each expert model processes inputs in which the modalities are kept isolated, and the coefficients are computed from these per-modality representations.

\begin{table*}[!ht]
\centering

\begin{minipage}{\textwidth}
\caption{Full results of the additional studies on probe prompt types for cross-modality tasks. We report accuracy across all datasets and their average per task group.}
\label{tab:ablation-prompt}
\centering
\small
\setlength{\tabcolsep}{2pt}
\renewcommand{\arraystretch}{1.2}
\resizebox{\textwidth}{!}{%
\footnotesize
\begin{tabular}{l ccc a cc a ccccc a ccc a}
\toprule
& \multicolumn{4}{c}{\textbf{Molecule-Protein Interaction}} & \multicolumn{3}{c}{\textbf{Molecule-Cell Interaction}} & \multicolumn{6}{c}{\textbf{CYP Inhibition}} & \multicolumn{4}{c}{\textbf{CYP Substrate}} \\
\cmidrule(l{2pt}r{2pt}){2-5} \cmidrule(l{2pt}r{2pt}){6-8} \cmidrule(l{2pt}r{2pt}){9-14} \cmidrule(l{2pt}r{2pt}){15-17}
Coefficient Type & BindingDB & BioSNAP & Human & \textit{Avg.} & DrugComb & GDSC2 & \textit{Avg.} & CYP1A2 & CYP2C19 & CYP2C9 & CYP2D6 & CYP3A4 & \textit{Avg.} & CYP2C9 & CYP2D6 & CYP3A4 & \textit{Avg.} \\
\midrule
\textbf{Separated Prompt} & \underline{65.3} & \underline{67.6} & \underline{61.9} & \underline{64.9} & \underline{77.1} & \underline{93.9} & \underline{85.5} & \underline{76.5}& \textbf{71.1}& \underline{71.2}& \underline{79.0}& \underline{69.6}& \underline{73.5}& \underline{59.4}& \underline{57.1}& \underline{61.9}& \underline{59.5}\\
\textbf{Concatenated Prompt} & \textbf{66.0} & \textbf{69.1} & \textbf{62.0} &  \textbf{65.7} & \textbf{80.7} & \textbf{94.1} & \textbf{87.4} & \textbf{77.4}& \underline{70.6}& \textbf{72.5}& \textbf{80.7}& \textbf{71.3}& \textbf{74.5}& \underline{64.2}& \textbf{60.9}& \textbf{60.5}& \textbf{61.9}\\
\bottomrule
\end{tabular}%
}
\end{minipage}

\vspace{0.1in} %

\begin{minipage}{\textwidth}
\caption{Full results of the additional studies on probe prompt types for single-modality tasks. We report accuracy and their average accuracy over all datasets.}
\label{tab:single_modality_prompt_ablation}
\centering
\setlength{\tabcolsep}{8pt}
\renewcommand{\arraystretch}{1.25}
\resizebox{0.45\textwidth}{!}{%
\begin{tabular}{l c c c >{\columncolor{gray!15}}c}
\toprule
& \textbf{Molecule} & \textbf{Protein} & \textbf{Cell} & \textit{Avg.} \\
\midrule
\textbf{Separated Prompt}             & \textbf{81.57} & \underline{80.83} & \underline{52.77} & \underline{71.72} \\
\textbf{Concatenated Prompt}             & \underline{77.64} & \textbf{86.51} & \textbf{61.49} & \textbf{75.21} \\
\bottomrule
\end{tabular}
}
\end{minipage}

\end{table*}

\paragraph{Analysis}
As summarized in Table~\ref{tab:ablation-prompt} and Table~\ref{tab:single_modality_prompt_ablation}, the concatenated prompt consistently outperforms the separated prompt across nearly all tasks. Not only on cross-modality tasks but also on single-modality tasks, the concatenated prompt achieves higher average accuracy than the separated prompt, with particularly large gains on the Protein and Cell modalities. This suggests that exposing all modality placeholders within a shared context allows each expert to produce coefficients that already encode cross-modal interaction, leading to more discriminative merging signals. In contrast, representations obtained from modality-isolated prompts lack the contrastive information across modalities, which in turn limits the modality-specific discriminativeness that the derived coefficients can provide. Based on this analysis, we adopt the concatenated prompt as our default probe setting in all main experiments.

\subsection{Probe Input Sensitivity}
\label{app:sec:probe_input_sensitivity}
We further validate the robustness of ES-Merging with respect to probe input construction. Specifically, we vary the size of probe inputs and compare the results with those obtained using another probe input set. As shown in Table~\ref{tab:ablation-sensitive}, ES-Merging with different probe input sizes generally outperforms the parameter-signal-based merging methods, demonstrating its robustness to the probe input size. Interestingly, increasing the probe input size does not bring a significant performance gain, while using a smaller size degrades the performance. This suggests that a size of 330 provides a favorable trade-off between performance and computational cost. On the other hand, the original probe input set shows a marginal performance gap compared to another set of the same size, indicating that ES-Merging is insensitive to the choice of probe inputs.

\begin{table}[!ht]
\centering
\small
\caption{Additional studies of probe dataset size and sampling for ES-Merging across Molecule-Protein and Molecule-Cell Interaction tasks.}
\vspace{0.1in}
\setlength{\tabcolsep}{8pt}
\renewcommand{\arraystretch}{1.2}
\begin{tabular}{lcccc} 
\toprule
Size & \textbf{Human} & \textbf{GDSC2} & \textbf{CYP2D6 Inh.} & \textbf{CYP2D6 Sub.}  \\
\midrule
50    & 61.0 & 91.2 & 78.7 & 59.4 \\
600    & 61.2 & 92.4 & 79.1 & 60.9 \\
330 (Another subset)  & 61.8 & 93.9 & 79.6 & 60.5 \\
330 & 62.0 & 94.1 & 80.7 & 60.9 \\
\bottomrule
\end{tabular}
\label{tab:ablation-sensitive}
\end{table}

\par\medskip
\noindent\begin{minipage}{\linewidth}
\centering
\captionof{table}{Mean accuracy (\%) and standard deviation of ES-Merging over three runs with independently sampled probe sets. The reported variability is due to probe sampling.}
\label{tab:probe_seed_stability}
\normalsize
\setlength{\tabcolsep}{8pt}
\renewcommand{\arraystretch}{1.15}
\begin{tabular}{lc}
\toprule
\textbf{Dataset} & \textbf{Mean $\pm$ Std.} \\
\midrule
Human & 61.8 $\pm$ 0.3 \\
GDSC2 & 93.9 $\pm$ 0.2 \\
CYP2D6 Inhibition & 80.2 $\pm$ 0.6 \\
CYP2D6 Substrate & 60.8 $\pm$ 0.2 \\
\bottomrule
\end{tabular}
\end{minipage}\par

\paragraph{Stability across Probe-Sampling Seeds}
As we use greedy decoding with temperature 0 and deterministic similarity-based few-shot retrieval, repeated inference with the same merged model and inputs produces identical predictions. Randomness arises when selecting probe inputs used to estimate the merging coefficients. To quantify this variability, we repeat ES-Merging three times using independently sampled probe sets and report the mean accuracy and standard deviation across runs in Table~\ref{tab:probe_seed_stability}. The standard deviations range from only 0.2 to 0.6 percentage points across the four evaluated datasets, indicating that ES-Merging remains stable under variations in probe sampling.

\subsection{Embedding Visualization}
\begin{figure*}[!ht]
    \centering
    \begin{minipage}[t]{0.32\linewidth}
        \centering
        \includegraphics[width=\linewidth]{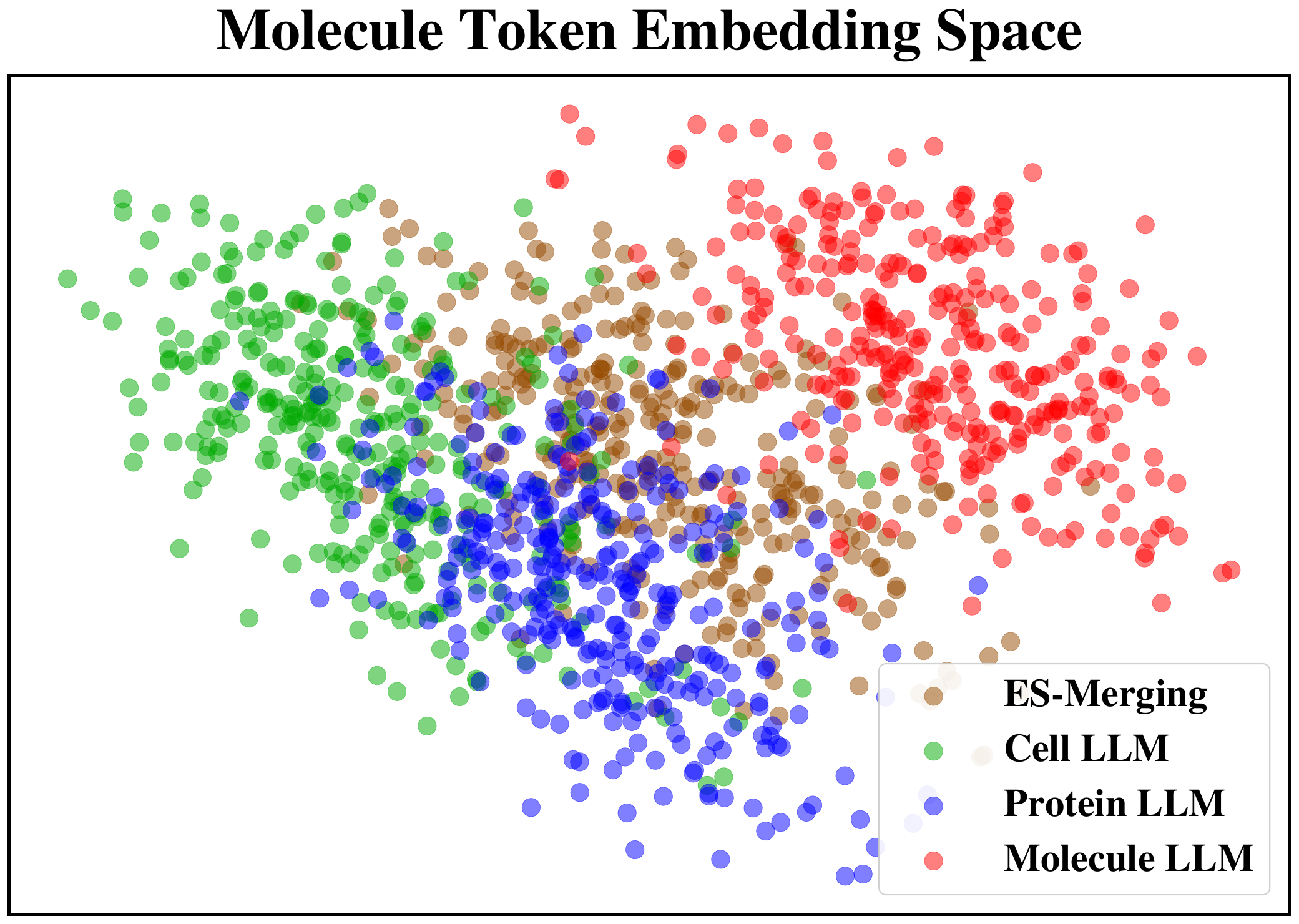}
        \vspace{-0.15in}
        \subcaption{Molecule token}
        \label{fig:mol_embedding_visualization_ours}
    \end{minipage}
    \hfill
    \begin{minipage}[t]{0.32\linewidth}
        \centering
        \includegraphics[width=\linewidth]{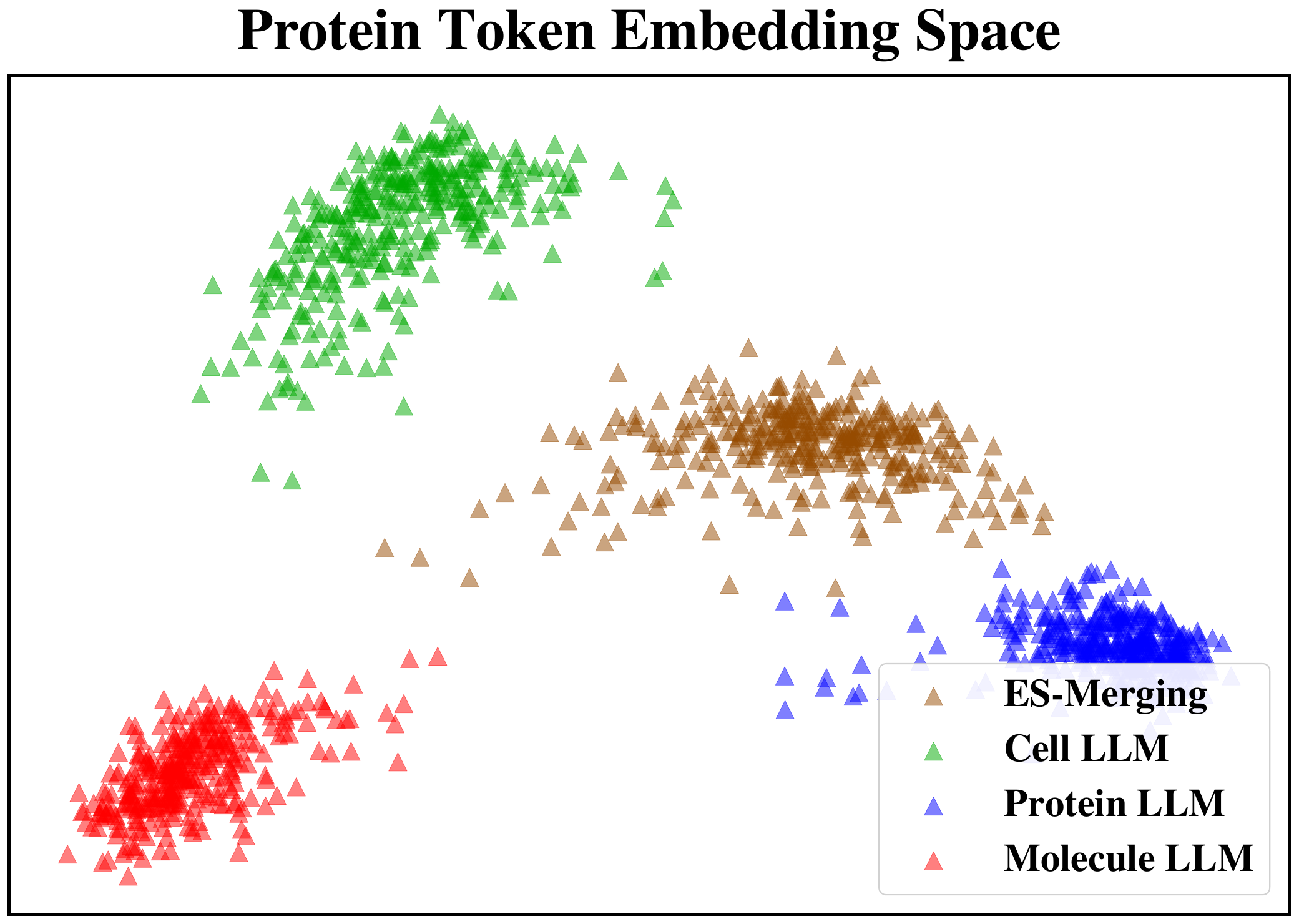}
        \vspace{-0.15in}
        \subcaption{Protein token}
        \label{fig:prot_embedding_visualization_ours}
    \end{minipage}
    \hfill
    \begin{minipage}[t]{0.32\linewidth}
        \centering
        \includegraphics[width=\linewidth]{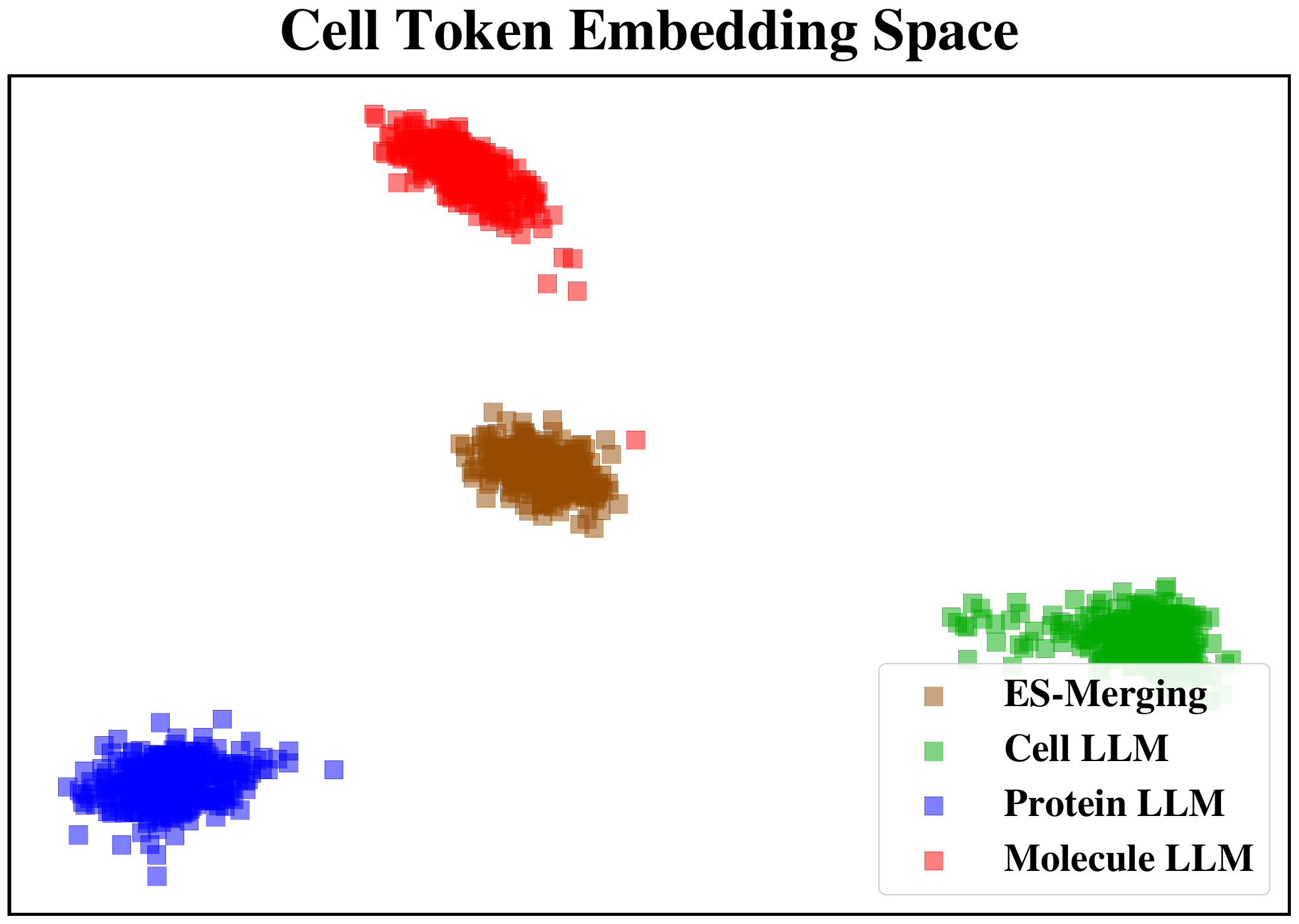}
        \vspace{-0.15in}
        \subcaption{Cell token}
        \label{fig:cell_embedding_visualization_ours}
    \end{minipage}
    
    \vspace{0.1in}
    
    \begin{minipage}[t]{0.32\linewidth}
        \centering
        \includegraphics[width=\linewidth]{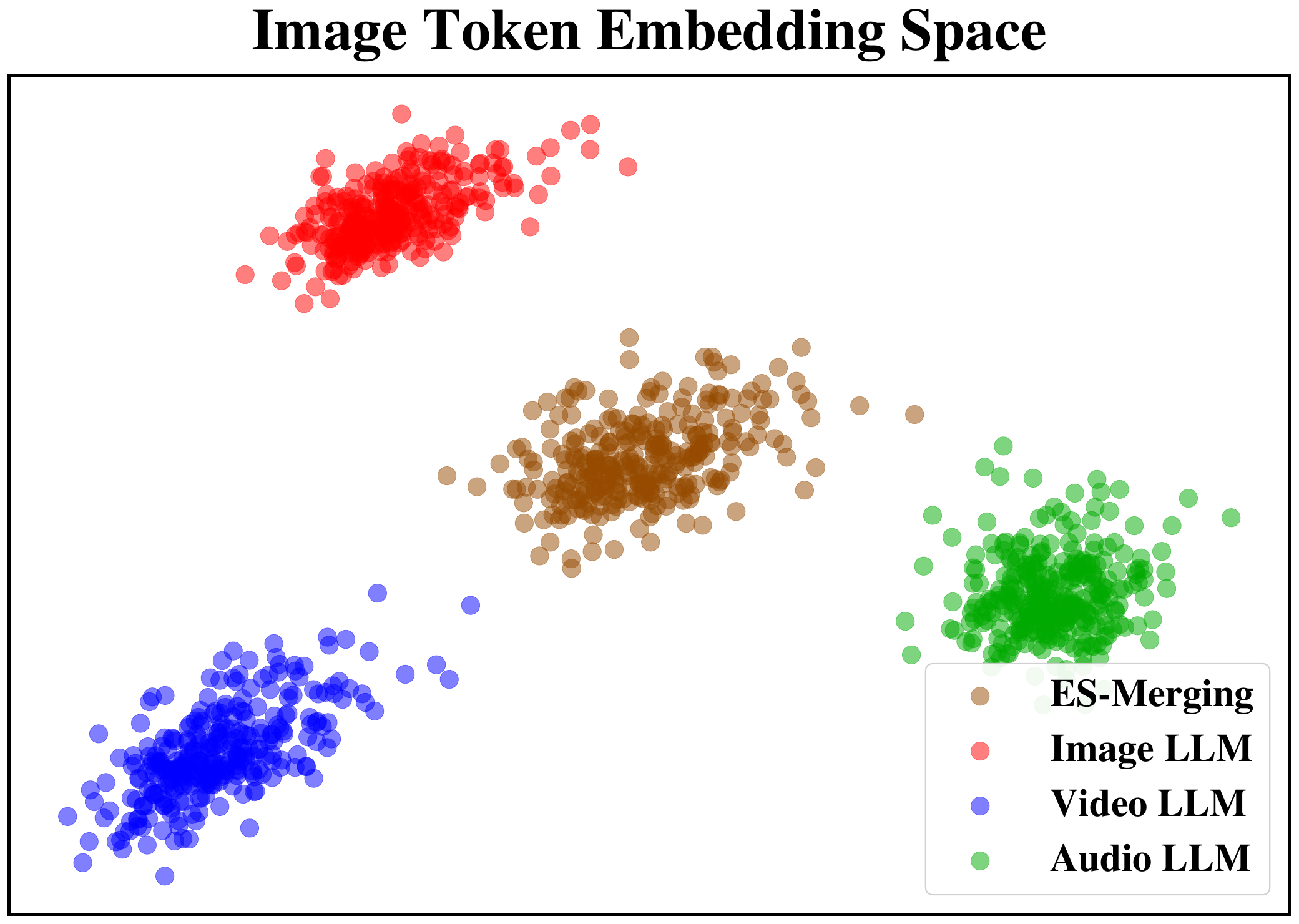}
        \vspace{-0.15in}
        \subcaption{Image token}
        \label{fig:image_embedding_visualization_ours}
    \end{minipage}
    \hfill
    \begin{minipage}[t]{0.32\linewidth}
        \centering
        \includegraphics[width=\linewidth]{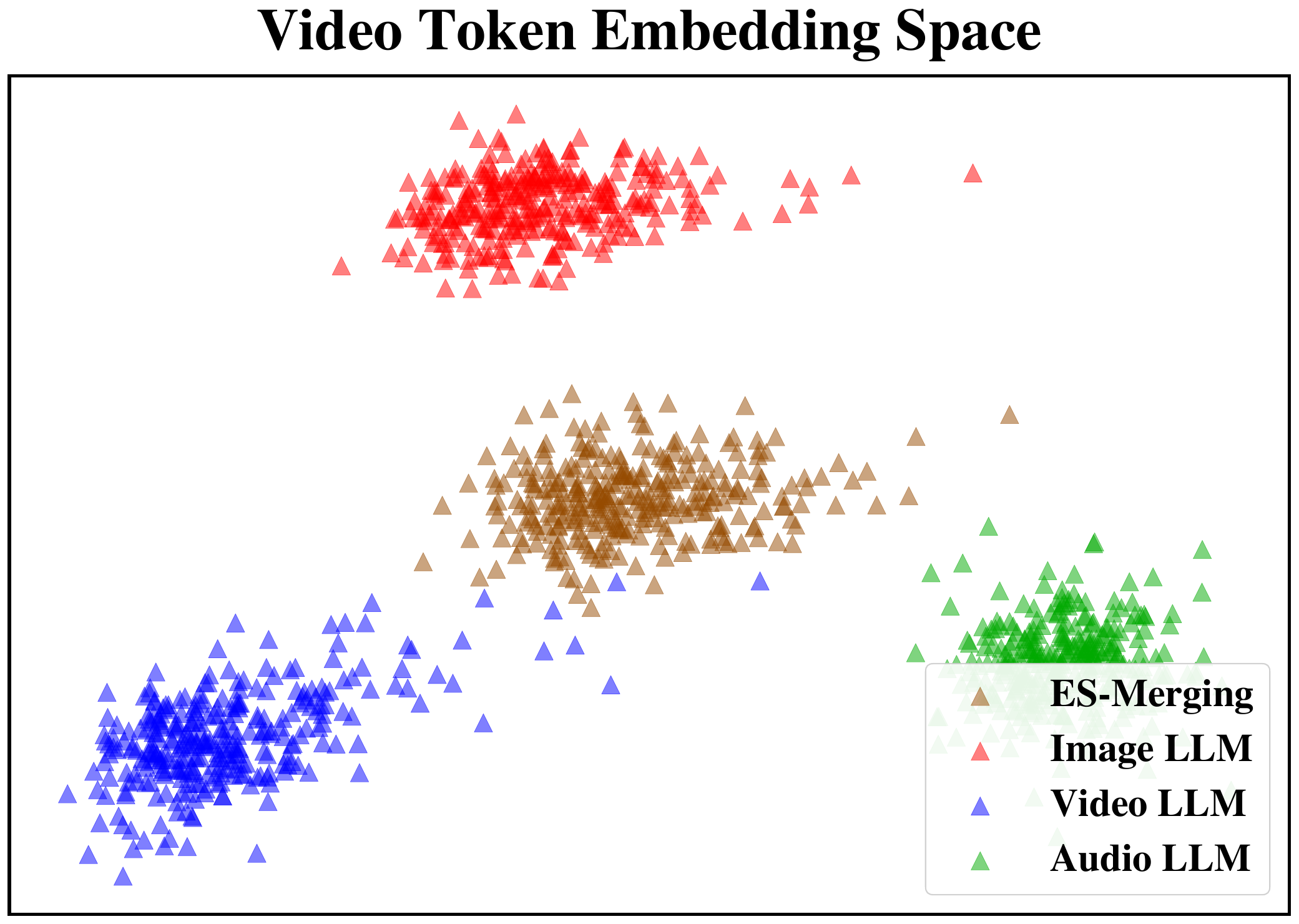}
        \vspace{-0.15in}
        \subcaption{Video token}
        \label{fig:video_embedding_visualization_ours}
    \end{minipage}
    \hfill
    \begin{minipage}[t]{0.32\linewidth}
        \centering
        \includegraphics[width=\linewidth]{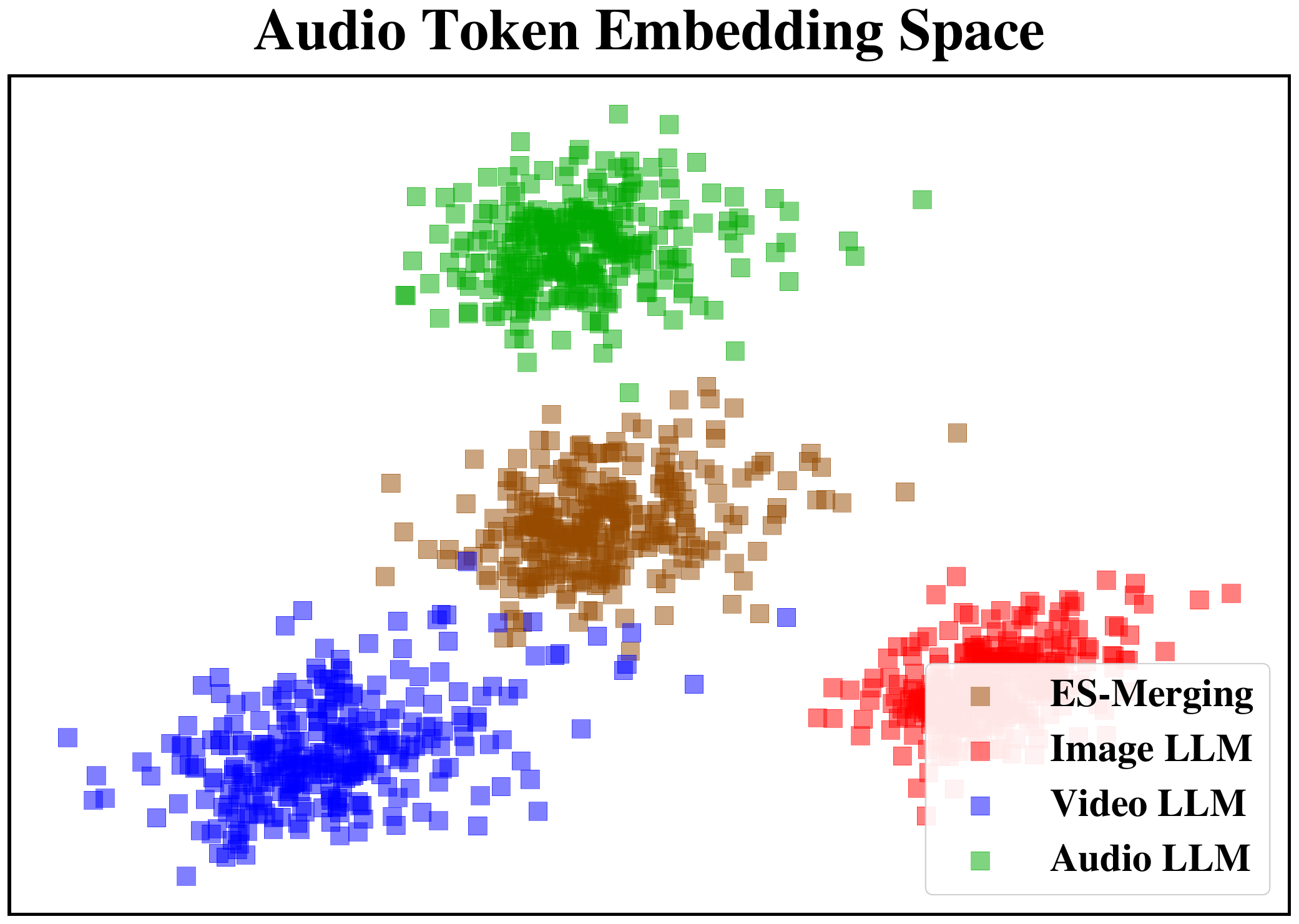}
        \vspace{-0.15in}
        \subcaption{Audio token}
        \label{fig:audio_embedding_visualization_ours}
    \end{minipage}
    \vspace{-0.05in}
    \caption{Embedding visualization of the last transformer block for each specialized LLM and the merged model with our method for (a) molecule, (b) protein, and (c) cell, (d) image, (e) video, and (f) audio tokens.}
    \label{fig:embedding_visualization_ours}
\end{figure*}
Figure~\ref{fig:embedding_visualization_ours} visualizes the embedding distributions of each specialist model and the model merged by ES-Merging at the last transformer block for each modality token. In each modality token space, the specialist models form distinct distributions, indicating that each model has learned modality-specific representations. The model merged by ES-Merging is positioned between the specialist distributions without being biased toward any particular specialist, while being relatively close to the distribution of the specialist corresponding to each modality token. This suggests that ES-Merging integrates the modality-specific knowledge of each specialist model in a balanced manner, while preserving the specialization for each modality. Notably, this pattern holds consistently across both biological and general domains, demonstrating that ES-Merging generalizes well across heterogeneous task settings rather than being tailored to a specific domain.

\subsection{Additional Evaluation on AVQA}
\label{app:sec:avqa}
\begin{table*}[t]
\centering
\caption{Performance comparison on AVQA across input-modality configurations. We report accuracy (\%); V, I, and A denote video, image, and audio, respectively. \textbf{Bold} and \underline{underline} indicate the best and second best performances.}
\vspace{-0.05in}
\normalsize
\setlength{\tabcolsep}{5pt}
\renewcommand{\arraystretch}{1.15}
\begin{tabular}{lccccccc}
\toprule
\textbf{Method} & \textbf{V} & \textbf{I} & \textbf{A} & \textbf{V + I} & \textbf{V + A} & \textbf{I + A} & \textbf{V + I + A} \\
\midrule
Avg. Merging & 62.20 & 73.32 & 58.48 & 66.40 & 67.87 & 75.86 & 69.06 \\
TIES-Merging~\citep{tiesmerging} & 43.38 & 69.99 & 55.78 & 43.73 & 46.62 & 71.43 & 52.16 \\
EMR-Merging~\citep{emrmerging} & \underline{63.90} & \underline{74.83} & \underline{59.12} & \underline{70.86} & \underline{71.82} & \underline{76.05} & \underline{71.18} \\
PCB-Merging~\citep{pcbmerging} & 40.59 & 68.55 & 54.69 & 39.66 & 43.12 & 70.34 & 48.61 \\
\midrule
\textbf{ES-Merging (Ours)} & \textbf{74.93} & \textbf{75.95} & \textbf{61.11} & \textbf{78.36} & \textbf{79.00} & \textbf{77.94} & \textbf{79.32} \\
\bottomrule
\end{tabular}
\label{tab:avqa_merging_prediction}
\end{table*}

To further examine generalization beyond musical scenarios, we evaluate the merged video-, image-, and audio-specialized model on AVQA~\citep{yang2022avqa}, which covers diverse real-world scenes, objects, actions, and sound-producing events. We compare with Average, TIES, EMR, and PCB merging under the same seven input-modality configurations as in MUSIC-AVQA. As shown in Table~\ref{tab:avqa_merging_prediction}, ES-Merging outperforms all evaluated baselines across all seven input configurations, showing consistent advantages in both single-modality and multimodal settings. For ES-Merging, all three two-modality combinations improve over their constituent single-modality inputs. It is also the only evaluated method to achieve its highest accuracy when all three modalities are provided, reaching 79.32\%. Together, these results suggest that ES-Merging can effectively use complementary modality information across diverse real-world scenes, extending its applicability beyond the musical scenarios evaluated in MUSIC-AVQA.

\section{Limitation and Societal Impacts\label{app:sec:limitation_and_societal_impact}}
\paragraph{Limitation}
In this work, we present ES-Merging, a novel MLLM merging framework that derives layer-wise and element-wise merging coefficients by leveraging embedding-space signals. Although ES-Merging achieves strong performance across cross-modal reasoning and single-modal knowledge preservation, it still requires an additional pre-merging stage to estimate embedding-space signals from probe inputs, introducing a one-time computational overhead compared to methods such as average merging. Nevertheless, this overhead is substantially lower than that of iterative optimization-based approaches, since ES-Merging estimates the merging coefficients once rather than repeatedly updating model parameters. Future work could explore more lightweight signal estimation strategies, such as reusing cached representations or estimating sparse coefficients.

\paragraph{Societal Impacts} We demonstrate that ES-Merging effectively integrates modality-specialized biological MLLMs by leveraging embedding space signals, enabling improved cross-modal effect prediction across diverse biological tasks. This capability has the potential to accelerate drug discovery by reducing reliance on costly and time-consuming wet-lab experiments. However, there is a risk that such capabilities could be maliciously exploited to identify harmful or toxic substances. We strongly hope that our method will be used solely for beneficial scientific purposes.

\vspace{0.2in}
\newpage
\newtcolorbox{promptbox}[1]{
  enhanced,
  colback=white, colframe=black!80,
  boxrule=0.5pt,
  arc=8pt, outer arc=8pt,
  top=4pt, bottom=3pt, left=6pt, right=6pt,
  boxsep=1pt,
  fonttitle=\bfseries\footnotesize,
  title={\centering #1},
  coltitle=black,
  colbacktitle=white,
  boxed title style={
    enhanced,
    colframe=black!80,
    colback=white,
    boxrule=0.5pt,
    arc=6pt, outer arc=6pt,
    top=2pt, bottom=2pt, left=6pt, right=6pt,
  },
  attach boxed title to top center={yshift=-\tcboxedtitleheight/2},
  before upper={\footnotesize\raggedright},
  segmentation hidden,
}

\newcommand{\instline}{%
  \par\vspace{2pt}%
  \noindent\hdashrule{\linewidth}{0.4pt}{2pt 1.5pt}%
  \par\vspace{2pt}%
}

\footnotesize
\captionsetup{width=0.96\linewidth, justification=raggedright, singlelinecheck=false}
\captionof{table}{In-context learning prompt templates for the Molecule-Protein Interaction and CYP prediction task. Angle-bracketed tokens are replaced with the corresponding encoder embeddings: \textcolor{blue}{\texttt{<protein>}} and \textcolor{red}{\texttt{<mol>}} for the target pair, and \textcolor{blue}{\texttt{<protein k>}} and \textcolor{red}{\texttt{<molecule k>}} for each of the $k$ few-shot examples. The \texttt{\{label k\}} is replaced with the ground-truth label string for each example pair in the few-shot context.}
\label{tab:prompt_DTI_CYP}

\vspace{2pt}
\begin{promptbox}{Prompt for Molecule-Protein Interaction Prediction}
\textbf{System}\\[1pt]
You are an expert specialized in drug discovery and molecular biology.
You will be given a protein and a molecule.
Your task is to \textbf{determine whether a given molecule interacts with a specific protein.}
\instline
\textbf{User}\\[1pt]
Determine whether the given molecule interacts with the protein by following the examples.
\vspace{1pt}
\begin{tcolorbox}[colback=gray!10,colframe=gray!30,boxrule=0.4pt,arc=3pt,
  left=3pt,right=3pt,top=2pt,bottom=2pt,
  before skip=2pt,after skip=2pt]
Examples: \\[1pt]
Example 1: \quad Protein: \textcolor{blue}{\texttt{<protein 1>}} \quad Molecule: \textcolor{red}{\texttt{<molecule 1>}} \quad Final answer: \texttt{\{label 1\}} \\
\makebox[\linewidth][c]{$\vdots$} \\
Example $k$: \quad Protein: \textcolor{blue}{\texttt{<protein k>}} \quad Molecule: \textcolor{red}{\texttt{<molecule k>}} \quad Final answer: \texttt{\{label k\}}
\end{tcolorbox}
\vspace{1pt}
\noindent Protein: \textcolor{blue}{\texttt{<protein>}} \quad Molecule: \textcolor{red}{\texttt{<mol>}} \par
Your final answer must be exactly one of: \textbf{`Final answer: Interacts' or `Final answer: Does not interact'.}
\end{promptbox}

\vspace{-2pt}
\begin{promptbox}{Prompt for \textbf{CYP Inhibition} Prediction}
\textbf{System}\\[1pt]
You are an expert specialized in drug discovery and molecular biology.
You will be given a protein and a molecule. Your task is to \textbf{determine whether a given molecule inhibits a specific protein.}
\instline
\textbf{User}\\[1pt]
Determine whether the given molecule inhibits the protein by following the examples.
\vspace{1pt}
\begin{tcolorbox}[colback=gray!10,colframe=gray!30,boxrule=0.4pt,arc=3pt,
  left=3pt,right=3pt,top=2pt,bottom=2pt,
  before skip=2pt,after skip=2pt]
Examples: \\[1pt]
Example 1: \quad Protein: \textcolor{blue}{\texttt{<protein 1>}} \quad Molecule: \textcolor{red}{\texttt{<molecule 1>}} \quad Final answer: \texttt{\{label 1\}} \\
\makebox[\linewidth][c]{$\vdots$} \\
Example $k$: \quad Protein: \textcolor{blue}{\texttt{<protein k>}} \quad Molecule: \textcolor{red}{\texttt{<molecule k>}} \quad Final answer: \texttt{\{label k\}}
\end{tcolorbox}
\vspace{1pt}
\noindent Protein: \textcolor{blue}{\texttt{<protein>}} \quad Molecule: \textcolor{red}{\texttt{<mol>}} \par
Your final answer must be exactly one of: \textbf{`Final answer: Inhibit' or `Final answer: Does not inhibit'.}
\end{promptbox}

\vspace{-2pt}
\begin{promptbox}{Prompt for \textbf{CYP Substrate} Prediction}
\textbf{System}\\[1pt]
You are an expert specialized in drug discovery and molecular biology.
You will be given a protein and a molecule. Your task is to \textbf{determine whether a given molecule is a substrate of a specific protein.}
\instline
\textbf{User}\\[1pt]
Determine whether the given molecule is a substrate of the protein by following the examples.
\vspace{1pt}
\begin{tcolorbox}[colback=gray!10,colframe=gray!30,boxrule=0.4pt,arc=3pt,
  left=3pt,right=3pt,top=2pt,bottom=2pt,
  before skip=2pt,after skip=2pt]
Examples: \\[1pt]
Example 1: \quad Protein: \textcolor{blue}{\texttt{<protein 1>}} \quad Molecule: \textcolor{red}{\texttt{<molecule 1>}} \quad Final answer: \texttt{\{label 1\}} \\
\makebox[\linewidth][c]{$\vdots$} \\
Example $k$: \quad Protein: \textcolor{blue}{\texttt{<protein k>}} \quad Molecule: \textcolor{red}{\texttt{<molecule k>}} \quad Final answer: \texttt{\{label k\}}
\end{tcolorbox}
\vspace{1pt}
\noindent Protein: \textcolor{blue}{\texttt{<protein>}} \quad Molecule: \textcolor{red}{\texttt{<mol>}} \par
Your final answer must be exactly one of: \textbf{`Final answer: Substrate' or `Final answer: Not a substrate'.}
\end{promptbox}
\newpage
\newtcolorbox[]{dciprompt}[1]{
  enhanced,
  colback=white,
  colframe=black!80,
  boxrule=0.5pt,
  arc=8pt, outer arc=8pt,
  left=8pt, right=8pt, top=8pt, bottom=8pt,
  boxsep=2pt,
  fonttitle=\bfseries\small,
  title={\centering #1},
  coltitle=black,
  colbacktitle=white,
  boxed title style={
    enhanced,
    colframe=black!80,
    colback=white,
    boxrule=0.5pt,
    arc=6pt, outer arc=6pt,
    top=3pt, bottom=3pt, left=8pt, right=8pt,
  },
  attach boxed title to top center={yshift=-\tcboxedtitleheight/2},
  before upper={\small\raggedright},
  segmentation hidden,
}

\begin{table*}[!ht]
\centering
\small
\captionsetup{width=0.96\linewidth, justification=raggedright, singlelinecheck=false}
\caption{In-context learning prompt template for the Molecule-Cell Interaction prediction task. Angle-bracketed tokens are replaced as follows: target molecule token \textcolor{red}{\texttt{<mol>}} and \textcolor{red}{\texttt{<molecule\,k>}} for each of the $k$ few-shot examples are substituted with the corresponding encoder embedding, while the gene tokens \textcolor{green!50!black}{\texttt{<gene>}} are replaced with the corresponding cell line string. The \texttt{\{label k\}} is replaced with the ground-truth label string for each example pair in the few-shot context.}
\label{tab:prompt-Cell}
\vspace{-6pt}
\begin{dciprompt}{Prompt for Molecule-Cell Interaction Prediction: GDSC2}
\textbf{System}\\[1pt]
You are an expert specialized in cancer pharmacogenomics and drug response prediction. You will be given an anticancer molecule and a cancer cell represented as a list of gene names ordered by expression, where the most highly expressed genes appear first in descending order. Your task is to predict whether the molecule will biologically suppress the cancer cell. \textbf{If the molecule is expected to suppress/inhibit cancer cell growth or viability, label it Sensitive, otherwise label it Resistant.}
\instline
\textbf{User}\\[1pt]
Determine whether the given cell is sensitive or resistant to the given molecule based on the ranked gene-expression list.\par
\smallskip
\begin{tcolorbox}[colback=gray!10,colframe=gray!30,boxrule=0.4pt,arc=3pt,
  left=4pt,right=4pt,top=4pt,bottom=4pt,before skip=0pt,after skip=4pt]
Examples:\par\smallskip
Example 1: \quad Cell: \textcolor{green!50!black}{\texttt{<gene 1>}} \quad Molecule: \textcolor{red}{\texttt{<molecule 1>}} \quad Final answer: \texttt{\{label 1\}} \\
\makebox[\linewidth][c]{$\vdots$} \\
Example $k$: \quad Cell: \textcolor{green!50!black}{\texttt{<gene k>}} \quad Molecule: \textcolor{red}{\texttt{<molecule k>}} \quad Final answer: \texttt{\{label k\}}
\end{tcolorbox}\par
\smallskip
Cell: \textcolor{green!50!black}{\texttt{<gene>}} \quad Molecule: \textcolor{red}{\texttt{<mol>}}\par
Your final answer should be formatted as either: \textbf{`Final answer: Sensitive' or `Final answer: Resistant'.}
\end{dciprompt}

\vspace{4pt}

\begin{dciprompt}{Prompt for Molecule-Cell Interaction Prediction: DrugComb}
\textbf{System}\\[1pt]
You are an expert specialized in cancer pharmacogenomics and anticancer drug-combination response prediction. You will be given two anticancer molecules and a cancer cell represented as a ranked list of gene names (highest expression first). \textbf{Your task is to predict the binary interaction outcome in terms of anticancer effect in that cell line.}\par
\smallskip
Labels:\par
Synergistic~-- the combination produces a stronger anticancer effect than expected,\par
Antagonistic~-- the combination produces a weaker anticancer effect than expected.
\instline
\textbf{User}\\[1pt]
Predict whether the drug pair is Synergistic or Antagonistic in the given cell line.\par
\smallskip
\begin{tcolorbox}[colback=gray!10,colframe=gray!30,boxrule=0.4pt,arc=3pt,
  left=4pt,right=4pt,top=4pt,bottom=4pt,before skip=0pt,after skip=4pt]
Examples:\par\smallskip
Example 1: \quad Cell: \textcolor{green!50!black}{\texttt{<gene 1>}} \quad Molecule1: \textcolor{red}{\texttt{<molecule1 1>}} \quad Molecule2: \textcolor{red}{\texttt{<molecule2 1>}} \\ Final answer: \texttt{\{label 1\}} \\
\makebox[\linewidth][c]{$\vdots$} \\
Example $k$: \quad Cell: \textcolor{green!50!black}{\texttt{<gene k>}} \quad Molecule1: \textcolor{red}{\texttt{<molecule k>}} \quad Molecule2: \textcolor{red}{\texttt{<molecule2 k>}} \\ Final answer: \texttt{\{label k\}}
\end{tcolorbox}\par
\smallskip
Cell: \textcolor{green!50!black}{\texttt{<gene>}} \quad Molecule1: \textcolor{red}{\texttt{<mol1>}} \quad Molecule2: \textcolor{red}{\texttt{<mol2>}}\par
Your final answer should be formatted as either: \textbf{`Final answer: Synergistic' or `Final answer: Antagonistic'.}
\end{dciprompt}
\end{table*}
\clearpage

\end{document}